\documentclass[10pt,twocolumn,letterpaper]{article}

\usepackage{iccv}
\usepackage{times}
\usepackage{epsfig}
\usepackage{graphicx}
\usepackage{amsmath}
\usepackage{amssymb}

\usepackage{customization}

\usepackage[toc,page]{appendix}
\newcommand{\stoptocwriting}{%
  \addtocontents{toc}{\protect\setcounter{tocdepth}{-5}}}
\newcommand{\resumetocwriting}{%
  \addtocontents{toc}{\protect\setcounter{tocdepth}{\arabic{tocdepth}}}}
\usepackage{tocloft}  

\usepackage[pagebackref=true,breaklinks=true,colorlinks,bookmarks=false]{hyperref}
\hypersetup{
    linkcolor=eq-fig-tab-color,  
    citecolor=ref-color,  
}
\iccvfinalcopy 

\def\iccvPaperID{1629} 

\newcommand\blfootnote[1]{%
  \begingroup
  \renewcommand\thefootnote{}\footnote{#1}%
  \addtocounter{footnote}{-1}%
  \endgroup
}

\begin{document}

\title{Spatio-Temporal Representation Factorization \\ for Video-based Person Re-Identification}

\author{Abhishek Aich$^{\star,2}$, Meng Zheng$^{1}$, Srikrishna Karanam$^{1}$, Terrence Chen$^{1}$,
\\ \vspace{-0.65\baselineskip} Amit K. Roy-Chowdhury$^{2}$, and Ziyan Wu$^{1}$\\
$^{1}$United Imaging Intelligence, Cambridge, MA, USA, $^{2}$University of California, Riverside, CA, USA\\
{\tt \small \{aaich001@, amitrc@ece.\}ucr.edu, \{first.last\}@uii-ai.com}
}

\maketitle

\begin{abstract}

Despite much recent progress in video-based person re-identification (re-ID), the current state-of-the-art still suffers from common real-world challenges such as appearance similarity among various people, occlusions, and frame misalignment. To alleviate these problems, we propose Spatio-Temporal Representation Factorization (STRF), a flexible new computational unit that can be used in conjunction with most existing 3D convolutional neural network architectures for re-ID. The key innovations of STRF over prior work include explicit pathways for learning discriminative temporal and spatial features, with each component further factorized to capture complementary person-specific appearance and motion information. Specifically, temporal factorization comprises two branches, one each for static features (e.g., the color of clothes) that do not change much over time, and dynamic features (e.g., walking patterns) that change over time. Further, spatial factorization also comprises two branches to learn both global (coarse segments) as well as local (finer segments) appearance features, with the local features particularly useful in cases of occlusion or spatial misalignment. These two factorization operations taken together result in a modular architecture for our parameter-wise light STRF unit that can be plugged in between any two 3D convolutional layers, resulting in an end-to-end learning framework. We empirically show that STRF improves performance of various existing baseline architectures while demonstrating new state-of-the-art results using standard person re-ID evaluation protocols on three benchmarks.

\end{abstract}

\captionsetup[figure]{list=no}
\captionsetup[table]{list=no}
\stoptocwriting
\section{Introduction}
\label{sec:intro}
\begin{figure}
    \centering
    \hspace*{-1.2em}
    \includegraphics[width=0.5\textwidth]{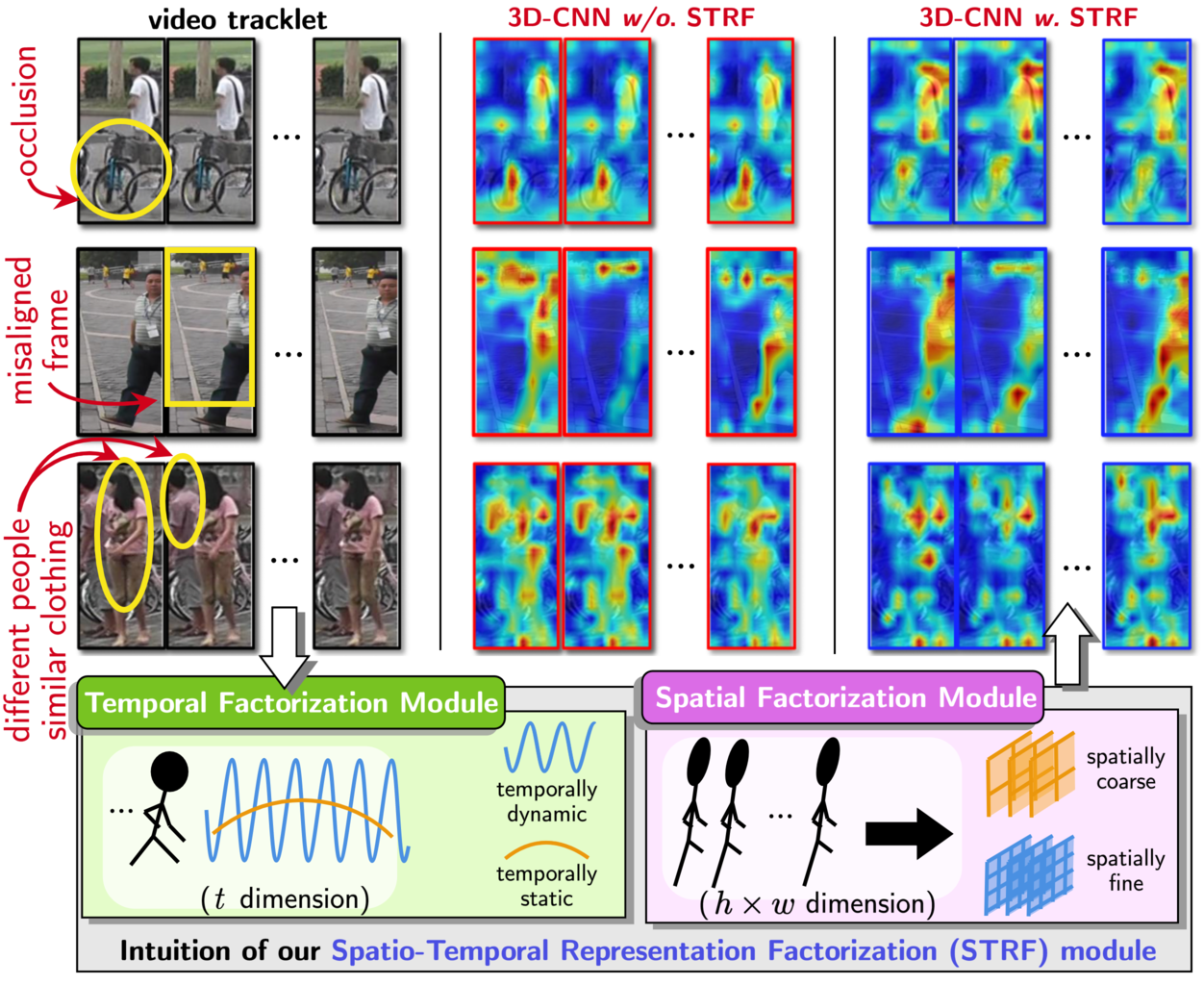}
    \caption{\textbf{Illustration of proposed concept and its efficacy.} We present the intuition behind our proposed Spatio-Temporal Representation Factorization (STRF) module designed to conquer common real-world re-ID system issues, e.g., similar-appearance identities, occlusions and misaligned frame. By capturing temporally static/dynamic and spatially coarse/fine information at different layers of a 3D-CNN, STRF produces robust discriminative representation to tackle these challenges as demonstrated here through attention maps of penultimate layer of feature extractor.}
    \vspace*{-\baselineskip}
    \label{fig:teaser}
\end{figure}


\blfootnote{$^\star$ This work was done during Abhishek Aich's internship with United Imaging Intelligence. Corresponding author: Srikrishna Karanam.}We consider the problem of video-based person re-IDentification (re-ID). Given a video tracklet of a person of interest, the task is to retrieve the closest match (which ideally should be the true match) among a gallery set of video tracklets. With numerous applications in security, surveillance, and forensics \cite{camps2016lab}, this problem has seen a dramatic increase in interest and various methodologies in the vision community \cite{gray2007evaluating,LOMO_XQDA_CVPR15, nam2017dual, LiHACNN_CVPR18, sunPCB_ECCV18, CASNimage_CVPR19, RAGAimage_CVPR2020, SGCSNimage_CVPR2020}. 


While there has been admirable progress in image-based re-ID as evidenced by recent quantitative results \cite{SGCSNimage_CVPR2020}, there are many challenges that still preclude the ubiquitous use of re-ID algorithms in real-world systems. One such issue is appearance similarity, where multiple people wear similar looking clothes (e.g., large conferences or public events with a strict dress code). Other challenging issues include occlusions and frame misalignment that are a direct consequence of large crowd flow densities (e.g., in airports just after flight arrival) and inter-camera viewpoint disparities. Having access to additional data, e.g., an extra temporal dimension like videos instead of 2D images, can help alleviate some of these issues by leveraging spatio-temporal data.


Video-based re-ID has seen much recent work \cite{li2018diversity, snippet_CVPR18, yan2020learning, TCLNet, yang2020spatial, zhang2020multi, gu2020AP3D, chentemporal} in part due to the availability of relevant large-scale video datasets \cite{zheng2016mars, wu2018exploit}. However, learning a spatio-temporal representation that can alleviate the issues noted above still remains a challenge. While advances in general 3D convolutional networks (3D-CNNs) provide reasonable baseline spatio-temporal features, existing re-ID techniques typically rely on specialized architectures \cite{yang2020spatial,hou2019vrstc,yan2020learning,zhang2020multi} that are inflexible to be used with these baseline models. 
Other lines of work are focused entirely on learning either temporal or spatial representations separately \cite{snippet_CVPR18, chentemporal, TCLNet}, overlooking the complementarity that both streams of information provide in challenging scenarios, e.g., distinguishing people wearing similar clothes.


To address the aforementioned issues, we present a flexible new computational unit called Spatio-Temporal Representation Factorization (STRF) module. Given a feature volume from a certain 3D convolutional layer in a baseline 3D-CNN model, STRF extracts complementary information along both spatial ($h\times w$) and temporal (time, $t$) dimensions. By design, the proposed STRF module can be inserted in an existing 3D-CNN model after any convolutional layer, introducing only $\sim$0.15 million learnable parameters per unit (for instance, this results in only a $\sim$1.73\% overall parameter increase with I3D \cite{carreira2017quo}), resulting in a flexible and parameter-wise economic framework that is end-to-end trainable. STRF comprises two modules, called \textit{temporal} feature factorization module (FFM) and \textit{spatial} feature factorization module, to process feature tensors. The design principles of these modules are motivated by certain observations in video tracklets, which we discuss next.

The intuition behind STRF is demonstrated in Figure~\ref{fig:teaser}. We begin with the factorization module in the temporal dimension. First, the overall or ``global" appearance of the person (e.g., color of clothes, skin, hair, etc) in a tracklet does not change (static) substantially over time. While one can argue these can change with illumination variations, we assume these variations are limited in a given camera view over a short period of time. Next, the walking patterns of a person may change over time, e.g., walking on a level surface \textit{vs}. climbing stairs (dynamic). Consequently, there are two possible information factorization strategies when processing feature maps: low-frequency (static) sampling and high-frequency (dynamic) sampling. Low-frequency sampling of feature maps results in capturing the ``slowly-moving" or approximately constant features, i.e., the appearance information. On the other hand, high-frequency sampling of feature maps results in capturing information that is more dynamically varying, i.e., walking patterns \cite{makihara2018gait}. The temporal factorization module results in capturing static and dynamic features across time, which is especially helpful in identifying different individuals with similar appearance (see last row video tracklet in Figure~\ref{fig:teaser}). 

The spatial factorization module, on the other hand, does the same low-frequency (which we call ``coarse") and high-frequency (which we call ``fine") sampling and processing as above, but along the spatial $h\times w$ dimensions. This is motivated by commonly occurring real-world issues such as occlusions and frame misalignment. Under these scenarios, the spatial FFM's high-frequency sampling and processing unit is able to capture more ``details" of the person of interest as opposed to the other entities that are the causes of occlusion, or other background information in the case of misalignment. To understand this better, observe the attention maps for top row video tracklet in Figure~\ref{fig:teaser}. The baseline model, without our proposed module, highlights mostly the bicycle regions in the feature maps, whereas by adding our module, the model is able to capture the person regions in the frames more comprehensively. Similarly, to cover cases where there are no occlusions or misalignment, the spatial FFM's low-frequency sampling and processing unit become responsible for capturing more slowly-varying or spatially global appearance information. This results in the spatial factorization module to capture two separate streams of spatial information for robust representations. 

To summarize, when multiple people in the gallery ``look alike" (e.g., same clothes), features from our temporal factorization branch help disambiguate (i.e., people may look alike but walk differently). On the other hand, with occlusion/clutter, our idea is to rely on ``local" features, which can be learned using our spatial branch. Our \textbf{main contributions} are as follows.
\begin{itemize}[leftmargin=*, topsep=0pt]
\setlength\itemsep{0.2em}
    \item We present a novel framework in video-based re-ID to learn discriminative 3D features by factorizing both temporal and spatial dimension of features into low-frequency (static/coarse) and high-frequency (dynamic/fine) components to tackle misalignment, occlusion, and similar appearance problems.
    \item To realize these factorization, we propose a flexible trainable unit with negligible computational overhead, called Spatio-Temporal Representation Factorization (STRF) module, that can be used in conjunction with any baseline 3D-CNN based re-ID architecture (see Figure \ref{fig:main-fig}).
    \item We conduct extensive experiments on multiple datasets to demonstrate how the proposed STRF module improves the performance of baseline architectures and also achieves state-of-the-art performance obtained by standard re-ID evaluation protocols (see Table \ref{tab:base_improve} and \ref{tab:SOTA}).
\end{itemize}
\section{Related Work}
In this section, we review some recent methods pertaining to video-based person re-ID, and later discuss 3D-CNNs as feature extractors for video re-ID tasks.

\begin{figure*}[t]
    \centering
    \includegraphics[width=\textwidth]{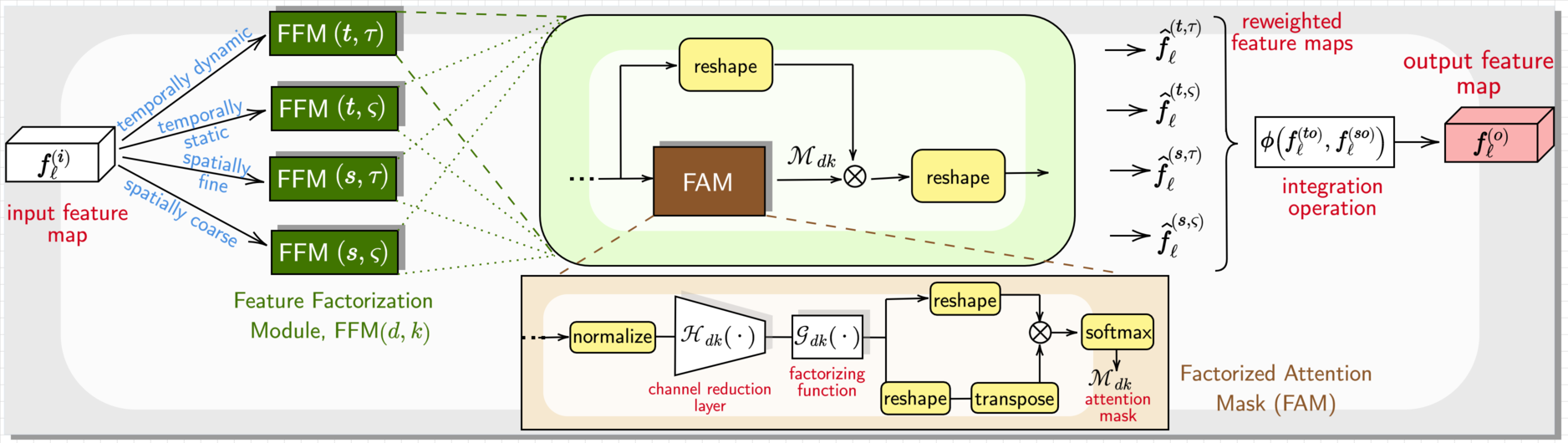}
    \caption{\textbf{Overview of STRF module}. Our proposed module contains four factorization units being applied on input feature volume at $\ell$th layer to extract static/coarse and dynamic/fine information and generate richer feature representation. Each unit is composed of a Feature Factorization Module aided by the proposed Factorized Attention Mask block.}
    \label{fig:main-fig}
    \vspace*{-\baselineskip}
\end{figure*}

\paragraph{Video-based re-ID.} Following the success in image-based re-ID \cite{LOMO_XQDA_CVPR15, nam2017dual, LiHACNN_CVPR18, sunPCB_ECCV18, CASNimage_CVPR19, RAGAimage_CVPR2020, SGCSNimage_CVPR2020, karanam2018systematic}, there has been much recent progress in video-based re-ID as well \cite{li2018diversity, xu2017jointly, zhou2017see, liao2018video, chentemporal, gu2020AP3D, TCLNet, gao2018revisiting}. For instance, \cite{yan2020learning} proposed multi-granular hypergraph learning framework which leveraged hierarchically divided feature maps at \textit{last} layer of feature encoder with different levels of granularities to capture spatial and temporal cues, treating both the spatial and temporal dimensions the same. Additionally, there have also been a class of methods \cite{TCLNet, li2019global} that perform feature modulation by \textit{expanding} the feature extractor with additional learning modules instead of processing just the last layer's output as in \cite{yan2020learning}. Different from all the above works, we focus on learning factorized (dynamic/static and coarse/fine) information in \textbf{both} spatial and temporal dimensions (see Figure \ref{fig:main-fig}). This leads to a flexible feature processing module that can be used anywhere in any 3D-CNN based re-ID architecture, leading to improved performance of various baseline 3D-CNN models (see Table \ref{tab:base_improve}). We provide a characteristic comparison of recent works in Table \ref{tab:compare_methods}.

\begin{table}[b]
\centering
\caption{\textbf{Characteristic comparison with state-of-the-art works.} We compare our STRF with few current state-of-the-art works. Unlike these methods, STRF uses factorized information from \textit{both} spatial (\textit{S}) and temporal (\textit{T}) dimensions, \textit{does not require} non-local operations, and is \textit{adaptable} to multiple baselines.}
\footnotesize{
\begin{tabular}{c|M{0.85cm}|M{0.85cm}|c|c}
\hlineB{2}
\multicolumn{1}{c|}{\multirow{2}{*}{\scriptsize \textbf{METHODS}}} & \multicolumn{2}{M{1.7cm}|}{\hspace{-1.6em}\notsotiny\textbf{FACTORIZATION}} & \multicolumn{1}{c|}{\multirow{2}{*}{\normalsize$\substack{\textbf{WITHOUT}\\[0.25em] \textbf{NON-LOCAL?}}$}} & \multicolumn{1}{c}{\multirow{2}{*}{\scriptsize\textbf{GENERIC?}}}\\ 
\cline{2-3}
 & \scriptsize{\textit{T}?} & \scriptsize{\textit{S}?} & & \\
\hlineB{1.5}
AP3D \cite{gu2020AP3D} & \ccross & \ccheck  & \ccross & \ccheck\\
\hline
MGH \cite{yan2020learning} & \ccheck & \raisebox{-0.2 em}{\ccheck} & \raisebox{-0.2 em}{\ccross} & \raisebox{-0.2 em}{\ccross} \\
\hline
AFA \cite{chentemporal} & \ccheck & \raisebox{-0.2 em}{\ccross}  & \raisebox{-0.2 em}{\ccheck} & \raisebox{-0.2 em}{\ccheck} \\
\hlineB{1.5}
\textbf{STRF} (Ours) & \ccheck & \ccheck  & \ccheck & \ccheck \\
\hlineB{2}
\end{tabular}}
\label{tab:compare_methods}
\end{table}

\paragraph{3D-CNN based Feature Extractor.} 3D-CNNs \cite{ji20123d} naturally process input videos to output spatio-temporal features, whereas 2D-CNNs need additional modules such as recurrent networks to extract temporal information. Given this advantage, 3D-CNNs are more suitable for video-related applications \cite{xu2017r, ji20123d, chen2018s3d, xie2018rethinking, feichtenhofer2019slowfast}, including video-based re-ID tasks \cite{liao2018video, 8999796, gu2020AP3D}. For example, \cite{1640938} introduced a two-stream model with the first branch comprised of 3D-CNNs and the other comprised of 2D-CNNs to extract temporal and spatial cues. In \cite{gu2020AP3D}, appearance-preserving 3D convolution (AP3D) was proposed to leverage the idea of image registration \cite{zitova2003image} to perform feature-level image alignment. While these methods demonstrated good results, they either required both 3D and 2D CNNs \cite{8999796}, or additional operations, e.g., non-local convolutions, to achieve best performances \cite{gu2020AP3D}, leading to parameter-wise bulky models. Furthermore, these methods do no explicitly exploit spatial cues of video tracklets. On the other hand, our proposed STRF method modifies the backbone feature encoder by means of a modular computational unit, does not require specialized modules such as recurrent networks or non-local operations, leading to only a minimal increase in learnable parameters while also demonstrating state-of-the-art performance on benchmark datasets.     


\section{Spatio-Temporal Factorization}
As noted in Section~\ref{sec:intro}, existing re-ID methods for learning video representations do not focus on the complementarity that is provided by the spatial and temporal dimensions. Specifically, we conjecture that the temporal dimension contains both static (e.g., appearance across time) as well as dynamic (e.g., walking patterns) content, whereas the spatial dimension comprises both fine (e.g., focus on details such as a person's legs that may be missed under occlusions) as well as coarse (e.g., overall global appearance) details. Consequently, we argue that all these features should be learned jointly in order to deal with unavoidable challenges such as appearance similarity, occlusions, and frame misalignment.

To address these issues, we introduce Spatio-Temporal Representation Factorization (STRF), a generic parameter-wise lightweight computational unit that can be inserted between convolutional layers in any 3D-CNN architecture for re-ID (note that by the term \textit{factorization}, we refer to the joint sampling and processing operations for discussion below). This modularity makes STRF particularly appealing for practical applications that may require customized architectures based on data distribution. Along with the performance improvements in baseline architectures (see Table \ref{tab:base_improve}), STRF also demonstrates superior utility of the proposed module over existing specialized architectures for learning spatio-temporal re-ID representations (see Table \ref{tab:SOTA}) \cite{yang2020spatial,8999796,hou2019vrstc,gu2020AP3D,yan2020learning}.


\paragraph{Notations.} Let $\bm{\mathcal{V}} = \begin{bmatrix}\bm{v}_1, \bm{v}_2, \cdots, \bm{v}_t\end{bmatrix} \in \mathbb{R}^{t\times h\times w}$ denote an input video tracklet comprising $t$ frames each of height $h$ and width $w$. Let $\mathcal{F}_{\bm{\theta}}(\cdot)$ denote the feature encoder of any baseline 3D-CNN (e.g., I3D ResNet-50 \cite{carreira2017quo}). Let $\bm{f}_\ell\in\mathbb{R}^{c_\ell \times t_\ell \times h_\ell \times w_\ell}$ be the feature tensor at the  $\ell$th layer of $\mathcal{F}_{\bm{\theta}}(\cdot)$, where $c_\ell, t_\ell, h_\ell,$ and $w_\ell$ indicate number of channels, number of frames, height, and width, respectively. Let the input and output feature volumes of our STRF module at the $\ell$th layer be $\bm{f}_\ell^{(\ip)}$ and $\bm{f}_\ell^{(\op)}$, respectively. Finally, let the static/coarse and dynamic/fine components be denoted with $\varsigma$ and $\tau$, respectively and subscript $\temp$ and $\spat$ denote the temporal and spatial dimension, respectively. We use $d\in\{\temp,\spat\}$ and $k\in\{\tau,\varsigma\}$ for compact notations. 

\subsection{Feature Factorization Module (FFM)}
Given $\bm{f}_\ell^{(\ip)}$, we propose to factorize this feature volume into four parts: static and dynamic content from temporal $t_\ell$ dimension, and coarse and fine detail from spatial $h_\ell \times w_\ell$ dimension. The intuition here is that the static content in the temporal dimension will capture ``what does not change over time", e.g., appearance such as color of clothes, and the dynamic content will capture ``what may change over time", e.g., walking patterns \cite{makihara2018gait}. Similarly, coarse details in the spatial dimension will capture overall global information in the current feature map (e.g., ``where is the person?") whereas fine detail helps address situations where the person of interest may be occluded by other entities, by capturing local context at different locations of the feature map.

Our motivations above are particularly relevant given existing 3D-CNN architectures for re-ID do not have explicit mechanisms to focus on features corresponding to the person of interest in cases such as occlusions, image misalignment, or people with similar clothing appearing together in same tracklet. Furthermore, such a factorization enables a 3D-CNN to \text{weight} features that are important for downstream matching and re-ID, e.g., the dynamic content along the temporal dimension is more important in cases where people wear similar clothes and can be distinguished only by their walking patterns. 

To realize this proposed factorization and feature re-weighting, STRF proposes to use four FFM modules, in which each FFM learns a different type of attention mask from \textbf{Factorized Attention Mask} (FAM) block (we discuss detailed architecture of FAM in next section) for either static/dynamic or coarse/fine content along the temporal and spatial dimensions respectively, and output refined feature volumes.
Specifically, given $\bm{f}_\ell^{(\ip)}$, we first reshape it into the feature volume $\widehat{\bm{f}}_\ell^{(\ip)}$ with size $c_\ell t_\ell  \times h_\ell w_\ell$ and then, use the FAM block to generate a factorized attention mask $\mathcal{M}_{dk}$. This mask is then used to compute a new feature volume as:
\begin{align}
    \widehat{\bm{f}}_\ell^{(dk)} = \widehat{\bm{f}}_\ell^{(\ip)}\mathcal{M}_{dk} \quad d \in \{\temp,\spat\}, k \in \{\tau, \varsigma\}
    \label{eq:out-eq}
\end{align}
STRF then integrates the four attention-weighted feature volumes $\{\widehat{\bm{f}}_\ell^{(\temp\tau)}, \widehat{\bm{f}}_\ell^{(\temp\varsigma)}, \widehat{\bm{f}}_\ell^{(\spat\tau)}, \widehat{\bm{f}}_\ell^{(\spat\varsigma)}\}$ to output a new feature volume which is then passed on to the subsequent convolutional layer. The output of this subsequent layer is then processed by the next instantiation of the STRF. This way, STRF provides a flexible computational unit that can be easily integrated with existing 3D-CNN architectures. Our proposed methodology is illustrated in Figure \ref{fig:main-fig} where one can note that the four individual factorization modules, FFM\big{(}$\temp, \tau$\big{)}, FFM\big{(}$\temp, \varsigma$\big{)}, FFM\big{(}$\spat, \tau$\big{)}, and FFM\big{(}$\spat, \varsigma$\big{)}, combine to produce an enhanced feature representation $\bm{f}_\ell^{(\op)}$ using their respective FAM blocks. We next discuss the factorization attention masks and each of these proposed individual FFM modules in more detail.

\subsection{Factorized Attention Masks (FAM) Block}

\begin{figure*}[!ht]
    \centering
    \includegraphics[width=\textwidth]{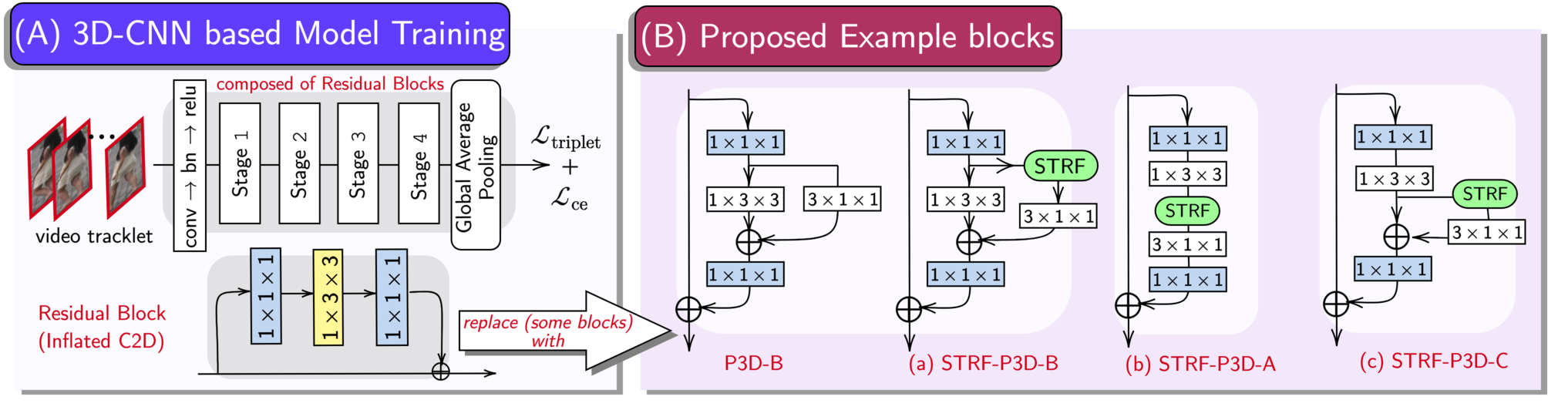}
    \caption{\textbf{Illustration of our 3D-CNN model training and examples of proposed blocks}. In our framework, we employ a 3D-CNN based model to learn discriminative features for input video tracklets (see Figure (\hyperref[fig:exp-fig]{A})). This model is built with inflated 2D-residual blocks with Stage \texttt{2} and Stage \texttt{3} replaced by our proposed STRF aided residual blocks (see Figure (\hyperref[fig:exp-fig]{B})).}
    \label{fig:exp-fig}
\end{figure*}
To realize the four-way factorization for the feature volume $\bm{f}_\ell^{(\ip)}$ discussed above, we define four functions below: 
\begin{alignat}{2}\label{eq:composite-function}
    \mathcal{T}^{k}_{d}\big{(}\bm{f}_\ell^{(\ip)}\big{)} &= \mathcal{G}_{dk}\bigg{(}\mathcal{H}_{dk}\big{(}\bm{f}_\ell^{(\ip)}\big{)}\bigg{)},\\
    \text{with,~~~~~~} d &\in \{\temp,\spat\}, k \in \{\tau, \varsigma\}\nonumber
\end{alignat}
where $\mathcal{G}_{dk}(\cdot)$ are the \textbf{factorizing functions}. Different for each FFM block, $\mathcal{G}_{dk}(\cdot)$ is designed using pooling functions to extract specific information after the input feature volume is passed through a channel reduction layer $\mathcal{H}_{dk}(\cdot):c_\ell \rightarrow \nicefrac{c_\ell}{n}$, where $\mathcal{H}_{dk}(\cdot)$ is a convolutional layer with $\nicefrac{c_\ell}{n}$ kernels of size 1. Following \cite{gu2020AP3D, jin2020style}, we set $n = 16$. With the output composite function $\mathcal{T}^{k}_{d}\big{(}\bm{f}_\ell^{(\ip)}\big{)}$ of size $\nicefrac{c_\ell}{n} t_\ell \times h_\ell w_\ell$ from \eqref{eq:composite-function}, we summarize input features by computing their variance matrix $\mathcal{C}_{dk}$ to obtain a representation of each point of $\mathcal{T}^{k}_{d}\big{(}\bm{f}_\ell^{(\ip)}\big{)}$ as:
\begin{align}
    \mathcal{C}_{dk} &= \kappa\mathcal{T}^{k}_{d}\big{(}\bm{f}_\ell^{(\ip)}\big{)}^\top\mathcal{T}^{k}_{d}\big{(}\bm{f}_\ell^{(\ip)}\big{)}
    \label{eq:cov-mat}
\end{align}
where $\top$ represents transpose operation. We set the temperature hyper-parameter $\kappa$ as 4 following \cite{gu2020AP3D, TCLNet}. Then, the factorizing mask is computed using the unnormalized sample covariance matrix as $\mathcal{M}_{dk}(q) =\sigma(\mathcal{C}_{dk})$, where $\sigma(\cdot)$ is the softmax function.
This factorized mask is employed in \eqref{eq:out-eq} to obtain the specific factorized representation of $\bm{f}_\ell^{(\ip)}$. Next, each factorization module is discussed in more detail.

\paragraph{Temporal Factorization Module, FFM\big{(}$\temp, \tau, \varsigma$\big{)}.} 
While methods for learning static and dynamic information have been presented in prior work \cite{feichtenhofer2019slowfast, aich2020non, tulyakov2018mocogan, gupta2021adavsr}, we take a modular approach to this problem, proposing computational units that can be applied at multiple layers of the base feature encoder. Instead of skipping frames as in \cite{feichtenhofer2019slowfast}, we define the following temporal factorizing functions:
\begin{equation}
\begin{gathered}
    \mathcal{G}_{\temp\tau} = \texttt{pool}\big{(}{r_{\temp\tau}, 1, 1}\big{)}, ~
    \mathcal{G}_{\temp\varsigma} = \texttt{pool}\big{(}{r_{\temp\varsigma}, 1, 1}\big{)}
    \label{eq:temp-fact-func}
\end{gathered}
\end{equation}
where $r_{\temp\varsigma} > r_{\temp\tau}$. These degenerate \texttt{pool} functions can be implemented using the \textit{max pooling} (denoted as $m$) and \textit{average pooling} (denoted as $a$) operations with their corresponding static temporal resolutions $r_{\temp\varsigma}$ and dynamic temporal resolutions $r_{\temp\tau}$. We also use suitable padding on $\mathcal{H}_{\temp\varsigma}(\bm{f}_\ell^{(\ip)})$ and $\mathcal{H}_{\temp\tau}(\bm{f}_\ell^{(\ip)})$ to maintain the same size between the input and output feature volumes. The intuition behind setting $r_{\temp\varsigma} > r_{\temp\tau}$ is to factorize features in time dimension to capture the information that does not vary much, whereas $r_{\temp\tau}$ helps in summarizing information that shows more variations. Capturing such static information with $\mathcal{G}_{\temp\varsigma}$ will aid in learning the global appearance features of the person that does not change much along the time dimension. On the other hand, $\mathcal{G}_{\temp\tau}$ captures dynamic information in the input feature volume, e.g., walking patterns of the person. Finally, the output of FFM\big{(}$\temp, \tau, \varsigma$\big{)} is defined as:
\begin{align}
    \bm{f}_\ell^{(\temp\op)} = \widehat{\bm{f}}_\ell^{(\temp\tau)} + \widehat{\bm{f}}_\ell^{(\temp\varsigma)}
\end{align}
where $\widehat{\bm{f}}_\ell^{(\temp\tau)}$ and $\widehat{\bm{f}}_\ell^{(\temp\varsigma)}$ are computed using \eqref{eq:out-eq}.

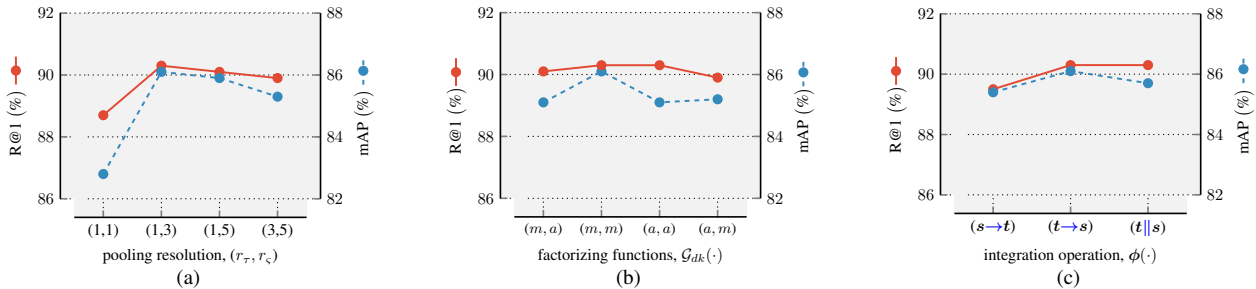
\begin{figure*}[!t]
     \hspace*{-0.3em}
     \captionsetup[subfigure]{aboveskip=-1pt,belowskip=-1pt}
     \begin{subfigure}[b]{0.33\textwidth}
          \centering
          \resizebox{0.9\linewidth}{!}{\pgfkeys{
   /pgf/number format/.cd, 
      set decimal separator={,{\!}},
      set thousands separator={}
}
\pgfplotsset{
   every axis/.append style = {
      line width = 1pt,
      tick style = {line width=1pt}
   }
}
\begin{tikzpicture}
    \pgfplotsset{
        height=4cm, width=5cm,
        grid = major,
        grid = both,
        grid style = {
        dash pattern = on 0.05mm off 1mm,
        line cap = round,
        black,
        line width = 0.75pt},
        scale only axis
    }
    \begin{axis}[
        xmin=0, xmax=8,
        xshift=-0.3cm,
        hide x axis,
        axis y line*=left,
        ymin=86, ymax=92,
        ylabel={R@1 \big{(}\%\big{)} \ref{pgfplots:r1_a}},
        axis background/.style={fill=gray!10}
    ]
    \end{axis}
    \begin{axis}[
        height=2cm, yshift=-0.4cm,
        xmin=0, xmax=8,
        ymin=86, ymax=92,
        axis x line*=bottom,
        hide y axis,
        xtick = {1, 3, 5, 7},
        xticklabels = {(1,1), (1,3), (1,5), (3,5)},
        xlabel={pooling resolution, ($r_\tau, r_\varsigma$)},
        axis background/.style={fill=gray!10}
    ]
    \end{axis}
    \begin{axis}[
        xmin=0, xmax=8,
        ymin=82.0, ymax=88.0,
        xshift=0.3cm,
        hide x axis,
        axis y line*=right,
        ylabel={mAP \big{(}\%\big{)} \ref{pgfplots:map_a}},
        axis background/.style={fill=gray!10}
    ]
    \end{axis}
    \begin{axis}[
        xmin=0, xmax=8,
        ymin=86, ymax=92,
        hide x axis,
        hide y axis,
    ]
        \addplot[very thick, r1-color, mark=*, mark options={solid, scale=1.25, fill=r1-color}] 
        coordinates {(1,88.70)(3,90.3)(5,90.1)(7,89.9)};
        \label{pgfplots:r1_a}
    \end{axis}
    \begin{axis}[
        xmin=0, xmax=8,
        ymin=82.0, ymax=88.0,
        hide x axis,
        hide y axis
    ]
        \addplot[dashed, very thick, map-color, 
        mark=*, mark options={solid, scale=1.25, fill=map-color}] coordinates {(1,82.8)(3,86.1)(5,85.9)(7,85.3)};
        \label{pgfplots:map_a}
    \end{axis}
    
\end{tikzpicture}}  
          \caption{}
          \label{fig:plot-A}
     \end{subfigure}
     \begin{subfigure}[b]{0.33\textwidth}
          \centering
          \resizebox{0.9\linewidth}{!}{\pgfkeys{
   /pgf/number format/.cd, 
      set decimal separator={,{\!}},
      set thousands separator={}
}
\pgfplotsset{
   every axis/.append style = {
      line width = 1pt,
      tick style = {line width=1pt}
   }
}
\begin{tikzpicture}
    \pgfplotsset{
        height=4cm, width=5cm,
        grid = major,
        grid = both,
        grid style = {
        dash pattern = on 0.05mm off 1mm,
        line cap = round,
        black,
        line width = 0.75pt},
        scale only axis
    }
    \begin{axis}[
        xmin=0, xmax=8,
        xshift=-0.3cm,
        width=2cm,
        hide x axis,
        axis y line*=left,
        ymin=86, ymax=92,
        ylabel={R@1 \big{(}\%\big{)}\ref{pgfplots:r1_b}},
        axis background/.style={fill=gray!10}
    ]
    \end{axis}
    \begin{axis}[
        height=2cm, yshift=-0.4cm,
        xmin=0, xmax=8,
        ymin=86, ymax=92,
        axis x line*=bottom,
        hide y axis,
        xtick = {1, 3, 5, 7},
        xticklabels = {($m, a$), ($m, m$), ($a, a$), ($a, m$)},
        xlabel={factorizing functions, $\mathcal{G}_{dk}(\cdot)$},
        xticklabel style = {font=\small},
        axis background/.style={fill=gray!10}
    ]
    \end{axis}
    \begin{axis}[
        xmin=0, xmax=8,
        ymin=82.0, ymax=88.0,
        xshift=0.3cm,
        hide x axis,
        axis y line*=right,
        ylabel={mAP \big{(}\%\big{)}\ref{pgfplots:map_b}},
        axis background/.style={fill=gray!10}
    ]
    \end{axis}
    \begin{axis}[
        xmin=0, xmax=8,
        ymin=82.0, ymax=88.0,
        hide x axis,
        hide y axis,
        ]
        \addplot[dashed, very thick, map-color, 
        mark=*, mark options={solid, scale=1.25, color=map-color, fill=map-color}] coordinates {(1,85.1)(3,86.1)(5,85.1)(7,85.2)};
        \label{pgfplots:map_b}
    \end{axis}
    \begin{axis}[
        xmin=0, xmax=8,
        ymin=86, ymax=92,
        hide x axis,
        hide y axis
        ]
        \addplot[very thick, r1-color, mark=*, mark options={solid, scale=1.25, color = r1-color, fill=r1-color}] coordinates {(1,90.1)(3,90.3)(5,90.3)(7,89.9)};
        \label{pgfplots:r1_b}
    \end{axis}
\end{tikzpicture}}  
          \caption{}
          \label{fig:plot-B}
     \end{subfigure}
     \begin{subfigure}[b]{0.33\textwidth}
          \centering
          \resizebox{0.9\linewidth}{!}{\pgfkeys{
   /pgf/number format/.cd, 
      set decimal separator={,{\!}},
      set thousands separator={}
}
\pgfplotsset{
   every axis/.append style = {
      line width = 1pt,
      tick style = {line width=1pt}
   }
}
\begin{tikzpicture}
    \pgfplotsset{
        height=39mm, width=5cm,
        grid = major,
        grid = both,
        grid style = {
        dash pattern = on 0.05mm off 1mm,
        line cap = round,
        black,
        line width = 0.75pt},
        scale only axis,
    }
    \begin{axis}[
        xmin=0, xmax=6,
        xshift=-0.3cm,
        hide x axis,
        axis y line*=left,
        ymin=86, ymax=92,
        ylabel={R@1 \big{(}\%\big{)}\ref{pgfplots:r1_c}},
        axis background/.style={fill=gray!10}
    ]
    \end{axis}
    \begin{axis}[
        height=2cm, yshift=-0.4cm,
        xmin=0, xmax=6,
        ymin=86, ymax=92,
        axis x line*=bottom,
        hide y axis,
        xtick = {1, 3, 5},
        xticklabels = {($\spat \textcolor{blue}{\bm{\rightarrow}} \temp$), ($\temp \textcolor{blue}{\bm{\rightarrow}} \spat$), ($\temp \textcolor{blue}{\bm{\Vert}} \spat$)},
        xlabel={integration operation, $\bm{\phi}(\cdot)$},
        axis background/.style={fill=gray!10}
    ]
    \end{axis}
    \begin{axis}[
        xmin=0, xmax=6,
        ymin=82.0, ymax=88.0,
        xshift=0.3cm,
        hide x axis,
        axis y line*=right,
        ylabel={mAP \big{(}\%\big{)} \ref{pgfplots:map_c}},
        axis background/.style={fill=gray!10}
    ]
    \end{axis}
    \begin{axis}[
        xmin=0, xmax=6,
        ymin=86, ymax=92,
        hide x axis,
        hide y axis
    ]
        \addplot[very thick, r1-color, mark=*, mark options={solid, scale=1.25, fill=r1-color}] coordinates {(1,89.5)(3,90.3)(5,90.3)};
        \label{pgfplots:r1_c}
    \end{axis}
    \begin{axis}[
        xmin=0, xmax=6,
        ymin=82.0, ymax=88.0,
        hide x axis,
        hide y axis
    ]
        \addplot[dashed, very thick, map-color, 
        mark=*, mark options={solid, scale=1.25, fill=map-color}] coordinates {(1,85.4)(3,86.1)(5,85.7)};
        \label{pgfplots:map_c}
    \end{axis}
    
\end{tikzpicture}}  
          \caption{}
          \label{fig:plot-C}
     \end{subfigure}
     \caption{\textbf{Analysis of different components of STRF}. (\hyperref[fig:plot-A]{a}) Each $(r_\varsigma, r_\varsigma)$ refers to spatially coarse resolutions $: (1, r_\varsigma, r_\varsigma)$, spatially fine resolutions $: (1, r_\tau, r_\tau)$, temporally static resolutions $: (r_\varsigma, 1, 1)$, temporally dynamic resolutions $: (r_\tau, 1, 1)$. Best results are obtained with $(r_\varsigma, r_\varsigma) = (1,3)$. (\hyperref[fig:plot-A]{b}) Performance of different combinations of factorizing functions $\mathcal{G}_{dk}(\cdot)$: Best results are obtained when all  $\mathcal{G}_{dk}(\cdot)$ are set as maxpooling function. (\hyperref[fig:plot-A]{c}) Performance of different integration operations $\bm{\phi}(\cdot)$: Best results are obtained when the spatial module is followed by the temporal module.}
    \vspace*{-\baselineskip}
 \end{figure*}

\paragraph{Spatial Factorization Modules, FFM\big{(}$\spat, \tau, \varsigma$\big{)}.}
Similar to the temporal dimension above, we factorize the feature volume along the spatial dimension as well, extracting coarse-level and fine-level information. The intuition here is that coarse-level information in the spatial dimension comprise global features of the person in the input frames that do not have much occlusion. For frames where the person is occluded or there is spatial misalignment, fine-level features capture the ``person-part" of the frame. To realize this, we define the following spatial factorizing functions:  
\begin{equation}
    \begin{gathered}
    \mathcal{G}_{\spat\tau} = \texttt{pool}\big{(}{1, r_{\spat\tau}, r_{\spat\tau}}\big{)}, ~
    \mathcal{G}_{\spat\varsigma} = \texttt{pool}\big{(}{1, r_{\spat\varsigma}, r_{\spat\varsigma}}\big{)}
    \label{eq:spat-fact-func}
    \end{gathered}
\end{equation}
where $r_{\spat\varsigma} > r_{\spat\tau}$ are the spatially coarse and fine resolution, respectively. As in FFM\big{(}$\temp, \tau, \varsigma$\big{)}, we use appropriate padding on $\mathcal{H}_{\spat\varsigma}(\bm{f}_\ell^{(\ip)})$ and $\mathcal{H}_{\spat\tau}(\bm{f}_\ell^{(\ip)})$ to maintain the same size between the input and output feature volumes. Finally, the output of FFM\big{(}$\spat, \tau, \varsigma$\big{)} is defined as: 
\begin{align}
    \bm{f}_\ell^{(\spat\op)} = \widehat{\bm{f}}_\ell^{(\spat\tau)} + \widehat{\bm{f}}_\ell^{(\spat\varsigma)}
\end{align}
where $\widehat{\bm{f}}_\ell^{(\spat\tau)}$ and $\widehat{\bm{f}}_\ell^{(\spat\varsigma)}$ are computed using \eqref{eq:out-eq}. Note that when the resolutions are set as 1 in \eqref{eq:spat-fact-func} and \eqref{eq:temp-fact-func}, the factorizing functions behave as identity mapping. In our experiments, we set $r_{\spat\tau} = r_{\temp\tau}$ and $r_{\spat\varsigma} = r_{\temp\varsigma}$ for simplicity. 

\paragraph{Integration and overall STRF output.} After computing $\bm{f}_\ell^{(\temp\op)}$ and $\bm{f}_\ell^{(\spat\op)}$ as discussed above, we provide two schemes to integrate them and generate the final feature volume output of our proposed STRF computational unit:
\begin{align}
    \bm{f}_\ell^{(\op)} &= \bm{\phi}\big{(}\bm{f}_\ell^{(\temp\op)}, \bm{f}_\ell^{(\spat\op)}\big{)} \quad\text{where,}~~\bm{\phi}\big{(}\cdot\big{)}\in\{\textcolor{blue}{\bm{\rightarrow}}, \textcolor{blue}{\bm{\Vert}}\}
    \label{eq:integration}
\end{align}
Here, $\textcolor{blue}{\bm{\rightarrow}}$ denotes using the temporal and spatial factorization modules in cascade, and $\textcolor{blue}{\bm{\Vert}}$ denotes using them in parallel. When in cascade, the input $\bm{f}_\ell^{(\ip)}$ is fed to both modules in sequence, \textit{i.e.} FFM\big{(}$\spat, \tau, \varsigma$\big{)} followed by FFM\big{(}$\temp, \tau, \varsigma$\big{)}, or \textit{vice-versa}. When in parallel, the outputs of FFM\big{(}$\spat, \tau, \varsigma$\big{)} and FFM\big{(}$\temp, \tau, \varsigma$\big{)} are simply added. In our experiments, we noticed only minor performance differences across these operations (see Figure \ref{fig:plot-C}). 

\paragraph{Learning Objective.} Any STRF-aided network can be trained in an end-to-end manner with following objective:
\begin{align}
    \mathcal{L} = \mathcal{L}_{\text{ce}} + \mathcal{L}_{\text{triplet}}    
\end{align}
where $\mathcal{L}_{\text{ce}}$ is the standard cross-entropy classification, $\mathcal{L}_{\text{triplet}}$ is the cosine distance based triplet loss with batch-hard mining \cite{hermans2017defense}, and $\mathcal{L}$ is the overall loss function. Note that our method demonstrates state-of-the-art results (see Table \ref{tab:SOTA}) without any re-ID tricks \cite{pathak2019video}, e.g. label smoothing \cite{szegedy2016rethinking}, in our learning objective.

\paragraph{How do we employ STRF?}
The problem of person re-ID benefited tremendously with introduction of residual blocks \cite{he2016deep, gou2018systematic}. With the backbone feature extractor as  inflated C2D (time dimension of kernel set to 1) residual network, we propose to enhance its feature representation learning paradigm by simply replacing residual blocks at different stages with different STRF-aided I3D or STRF-aided Pseudo-3D (P3D) \cite{qiu2017learning} residual blocks (see Figure \hyperref[fig:exp-fig]{2(B)}). To convert P3D residual blocks to their STRF-P3D forms, we add the STRF module with the convolutional layer of kernel size $3\times 1\times 1$ demonstrating the generic ability of the proposed unit. We have empirically analyzed and discussed this choice of location in the supplementary material. Moreover, a single STRF module introduces only minimal extra-parameters which makes it parameter-wise lightweight but performance-wise beneficial (see Table \ref{tab:base_improve}). 


\subsection{Discussion}
\paragraph{FAM $\bm{vs}$ Channel Attention (CA).} We note that there are substantial differences between FAM and the popular CA strategy \cite{zhang2018image, dai2019second, gupta2020alanet}. Unlike CA that has one \textit{global} feature pooling layer, i.e., no separate spatial and temporal operations, FAM has \textbf{four} pooling functions $\mathcal{G}_{dk}(\cdot)$, defined in \eqref{eq:temp-fact-func} and \eqref{eq:spat-fact-func}. This captures both spatial and temporal feature dependencies without \textit{any} new learning parameters. In fact, with $r_\varsigma$ and $r_\tau$ set to same size of the input feature maps, CA can be considered to be a special case of FAM.
\paragraph{FFM $\bm{vs}$ Non-Local (NL).} Unlike the popular NL module \cite{wang2018non} where there is \textit{no} factorization, FFM factorizes $f^{(i)}$ into its constituent spatio/temporal factors. The appropriate weighting of $f^{(i)}$ with these factors to obtain $f^{(o)}$ is automatically learned with FAM, making the proposed design different from NL and more suitable for re-ID. For additional empirical substantiation, using the P3DC architecture on the MARS dataset \cite{zheng2016mars}, the NL module gives 84.8\% mAP and 89.9\% R@1, whereas STRF gives 86.1\% mAP and 90.3\% R@1. Further, STRF only adds an additional $\sim$0.5 million parameters (w.r.t. the baseline) as opposed to NL's $\sim$5 million additional parameters, demonstrating better compute efficiency.\\ 

Please see supplementary material for additional insights and discussions on our proposed STRF module.

\section{Experimentation}
\label{sec:results}

\paragraph{Datasets, implementation details, and evaluation metrics.} We conduct extensive experiments on standard publicly available video-based person re-ID datasets, including MARS \cite{zheng2016mars}, DukeMTMC-VideoReID \cite{wu2018exploit}, and iLIDS-VID
\cite{wang2014person}. For evaluation, we use the value of the cumulative matching characteristic curve at rank-1 (R@1), and mean average precision (mAP) \cite{zheng2015scalable}. See supplementary material for full implementation details. 

\subsection{Improvement over Baselines}
\paragraph{Quantitative analysis.} We build a model with inflated 2D convolutions in ResNet50 (temporal kernel size set to 1) architecture. We then replace stage \texttt{2} and stage \texttt{3} (See Table \ref{tab:stage_add}) with four residual blocks \textbf{I3D} (temporal kernel size set to 3) and three pseudo-3D residual blocks \textbf{P3D-A}, \textbf{P3D-B} and \textbf{P3D-C} to create four baselines. For comparative evaluation, we replace these I3D and P3D residual blocks with \textbf{STRF-I3D}, \textbf{STRF-P3DA}, \textbf{STRF-P3DB} and \textbf{STRF-P3DC} residual blocks respectively and summarize the results in Table \ref{tab:base_improve}. One can clearly note that the STRF-aided models give improved performance (at least 2.5\% mAP increment for P3D baselines and about 0.5\% mAP increment for I3D baseline on MARS), with the best performance achieved with STRF-P3DC. Similar trends can be observed on the  DukeMTMC-VideoReID as well. Furthermore, when compared to the number of baseline model parameters (denoted in Table~\ref{tab:base_improve} as \textbf{P}(M) on MARS in the millions of parameters), the number of new parameters introduced by our proposed module is only 0.05 million more compared with I3D or P3D models, suggesting it does not add any substantial computational overhead. This also demonstrates that STRF can improve performance of \textit{diverse} architectures. For all subsequent experiments, we report results with STRF-P3DC following its best performance from Table~\ref{tab:base_improve}. 

\begin{table}[ht]
    \centering
    \caption{\textbf{Baseline improvement.} STRF consistently improves the performance of baseline models. \textbf{P}(M) is model size in millions.}
    \footnotesize{
    \begin{tabular}{>{\centering\arraybackslash}p{1cm}|>{\centering\arraybackslash}p{0.5cm}|>{\centering\arraybackslash}p{1.0cm}|>{\columncolor{col-color!30}}>{\centering\arraybackslash}p{1.0cm}||>{\centering\arraybackslash}p{1.0cm}|>{\columncolor{col-color!30}}>{\centering\arraybackslash}p{1.0cm}}
    \hlineB{2}
    \multicolumn{1}{c|}{\multirow{3}{*}{\textbf{MODEL}}} &
    \multicolumn{1}{c|}{\multirow{3}{*}{\textbf{P}(M)}} & \multicolumn{4}{c}{\textbf{DATASETS}}\\
    \hhline{|>{\arrayrulecolor{white}}->{\arrayrulecolor{white}}->{\arrayrulecolor{black}}|-|-|-|-}
    & & \multicolumn{2}{c||}{\scriptsize{\textbf{MARS} \cite{zheng2016mars}}}
    & \multicolumn{2}{c}{\scriptsize{\textbf{DukeMTMC} \cite{wu2018exploit}}}\\
    \hhline{|>{\arrayrulecolor{white}}->{\arrayrulecolor{white}}->{\arrayrulecolor{black}}|-|-|-|-}
    & & \hspace*{-0.15em}\scriptsize mAP \big{(}\%\big{)} & \hspace*{-0.15em}\scriptsize R@1 \big{(}\%\big{)} & \hspace*{-0.15em}\scriptsize mAP \big{(}\%\big{)} & \hspace*{-0.15em}\scriptsize R@1 \big{(}\%\big{)} \\
    \hlineB{1.5}
    I3D
    & 28.92 & 82.70    & 88.50    & 95.20    & 95.40    \\
    \hline
    + \textbf{STRF} 
    & 28.97 & \textbf{83.10} & \textbf{88.70   } & \textbf{95.20} & \textbf{95.90} \\
    \hlineB{1.5}
    P3DA 
    & 25.48 & 83.20    & 88.90    & 95.00    & 95.00    \\
    \hline
    + \textbf{STRF}  
    & 25.53 & \textbf{85.40   } & \textbf{89.80   } & \textbf{95.60   } & \textbf{96.00   } \\
    \hlineB{1.5}
    P3DB 
    & 25.48 & 83.00    & 88.80    & 95.40    &  95.30    \\
    \hline
    + \textbf{STRF} 
    & 25.53 & \textbf{85.60   } & \textbf{90.30   } & \textbf{96.40   } & \textbf{97.40   } \\
    \hlineB{1.5}
    P3DC 
    & 25.48 & 83.10    & 88.50    & 95.30    & 95.30    \\
    \hline
    + \textbf{STRF}  
    & 25.53 & \textbf{86.10} & \textbf{90.30} & \textbf{96.20} & \textbf{97.20} \\
    \hlineB{2}
    \end{tabular}
    }
    \label{tab:base_improve}
    \vspace*{-\baselineskip}
\end{table}

\begin{figure}[t]
    \centering
    \vspace*{-\baselineskip}
    \hspace*{-1.1em}
    \includegraphics[width=1.05\columnwidth]{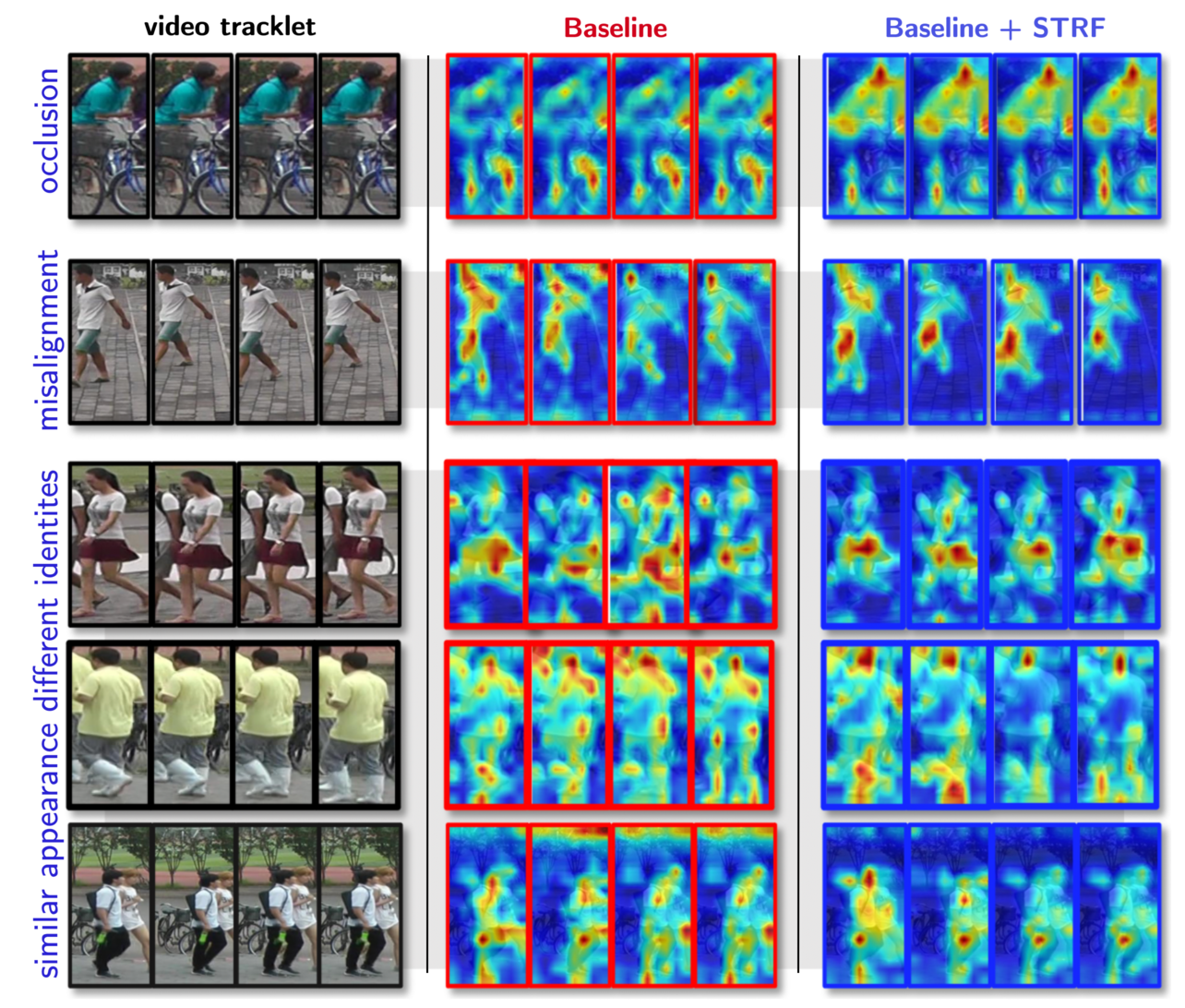}
    \caption{\textbf{Attention map visualizations.} STRF helps baseline models extract more discriminative features.}
    \label{fig:att-map-fig}
    \vspace*{-1\baselineskip}
\end{figure}

\paragraph{Qualitative Analysis.} To qualitatively demonstrate STRF's impact, we visualize feature maps of challenging videos (e.g., occlusions, misalignment) using off-the-shelf techniques \cite{zagoruyko2016paying,gu2020AP3D} in Figure \ref{fig:att-map-fig}. Note that STRF helps focus more clearly on the person of interest (e.g., under ``occlusion", unlike the baseline, STRF is able to more clearly distinguish between person's foreground and occlusion regions). Please see supplementary material for more qualitative and attention map results. 
\begin{table*}[!ht]
    \centering
    \caption{\textbf{Comparison with the state-of-the-art.} STRF gives state-of-the-art performance on all datasets (best results in \textcolor{best}{\textbf{red}}, second best in \textcolor{sec-best}{\textbf{blue}}, and third best results in \textcolor{third-best}{\textbf{green}}.)}
    \footnotesize{%
    \begin{tabular}{M{3cm}|M{2cm}|M{1.5cm}|>{\columncolor{col-color!30}}M{1.5cm}||M{1.5cm}|>{\columncolor{col-color!30}}M{1.5cm}||>{\columncolor{col-color!30}}M{1.85cm}}
    \hlineB{2}
    \multicolumn{1}{c|}{\multirow{3}{*}{\textbf{METHODS}}} &
    \multicolumn{1}{c|}{\multirow{3}{*}{\textbf{VENUE}}} &
    \multicolumn{5}{c}{\textbf{DATASETS}}\\
    \cline{3-7}
    &
    & \multicolumn{2}{c||}{\footnotesize{\textbf{MARS} \cite{zheng2016mars}}}
    & \multicolumn{2}{c||}{\footnotesize{\textbf{DukeMTMC} \cite{wu2018exploit}}} 
    & \multicolumn{1}{c}{\footnotesize{\textbf{iLiDS-VID} \cite{wang2014person}}}\\
    \hhline{|>{\arrayrulecolor{white}}->{\arrayrulecolor{white}}->{\arrayrulecolor{black}}|-|-|-|-|-}
    & & mAP \big{(}\%\big{)} & R@1 \big{(}\%\big{)} & mAP \big{(}\%\big{)} & R@1 \big{(}\%\big{)} & R@1 \big{(}\%\big{)}\\
    \hlineB{1.5}
    ADFD \cite{zhao2019attribute} & CVPR 2019 & 78.20~ & 87.00~ & -- & -- & 86.30~ \\
    \hline
    VRSTC \cite{hou2019vrstc} & CVPR 2019 & 82.30~ & 88.50~ & 93.50~ & 95.00~ & 86.30~ \\
    \hline
    COSAM \cite{subramaniam2019co} & ICCV 2019 & 79.90~ & 84.90~ & 94.10~ & 95.40~ & 79.60~ \\
    \hline
    GLTR \cite{li2019global} & ICCV 2019 & 78.50~ & 87.00~ & 93.74~ & 96.29~ & 86.00~ \\
    \hline
    MGH \cite{yan2020learning} & CVPR 2020 & \textcolor{third-best}{\textbf{85.80~}} & 90.00~ & -- & -- & 85.60~ \\
    \hline
    STGCN \cite{yang2020spatial} & CVPR 2020 & 83.70~ & 89.95~ & 95.70~ & \textcolor{sec-best}{\textbf{97.29~}} & -- \\
    \hline
    MG-RAFA \cite{zhang2020multi} & CVPR 2020 & \textcolor{sec-best}{\textbf{85.90~}} & 88.80~ & -- & -- & 88.60~ \\
    \hline
    TACAN \cite{li2020temporal} & WACV 2020 & 84.00~ & 89.10~ & 95.40~ & 96.20~ & \textcolor{sec-best}{\textbf{88.90~}} \\
    \hline
    M3D \cite{8999796} & TPAMI 2020 & 79.46~ & 88.63~ & 93.67~ & 95.49~ & 86.67~ \\
    \hline
    AFA \cite{chentemporal} & ECCV 2020 & 82.90~ & \textcolor{third-best}{\textbf{90.20~}} & 95.40~ & \textcolor{third-best}{\textbf{97.20~}} & 88.50~ \\
    \hline 
    AP3D \cite{gu2020AP3D} & ECCV 2020
    & 85.60~ & \textcolor{best}{\textbf{90.70~}} & \textcolor{third-best}{\textbf{96.10~}} & \textcolor{third-best}{\textbf{97.20~}} & \textcolor{third-best}{\textbf{88.70~}} \\
    \hline
    TCLNet \cite{TCLNet} & ECCV 2020
    & 85.10~ & 89.80~ & \textcolor{sec-best}{\textbf{96.20~}} & 96.90~ & 86.60~ \\
    \hlineB{1.5}
    \textbf{STRF} & \textbf{Ours}
    & \textcolor{best}{\textbf{86.10~}} & \textcolor{sec-best}{\textbf{90.30~}} & \textcolor{best}{\textbf{96.40~}} & \textcolor{best}{\textbf{97.40~}} & \textcolor{best}{\textbf{89.30~}} \\
    \hlineB{2}
    \end{tabular}
    }
\label{tab:SOTA}
\end{table*}
\begin{table}[b]
    \centering
    \caption{\textbf{Contribution of each factorization module.} All four STRF modules FFM$(\temp,\tau)$, FFM$(\temp,\varsigma)$ FFM$(\spat,\tau)$, and FFM$(\spat,\varsigma)$ show improvement individually and collectively with the P3DC baseline on MARS \cite{zheng2016mars}.}
    \scriptsize{
    \begin{tabular}{M{0.75cm}||>{\centering\arraybackslash}p{0.5cm}|>{\centering\arraybackslash}p{0.5cm}|>{\centering\arraybackslash}p{0.5cm}|>{\centering\arraybackslash}p{0.5cm}||>{\centering\arraybackslash}p{1.15cm}|>{\columncolor{col-color!30}}>{\centering\arraybackslash}p{1.15cm}}
    
    \hlineB{2}
    \multirow{2}{*}{\textbf{MODEL}} &
    \multicolumn{4}{c||}{\textbf{MODULES}} & \multicolumn{2}{c}{\textbf{MARS} \cite{zheng2016mars}} \\
    \hhline{|>{\arrayrulecolor{white}}->{\arrayrulecolor{black}}|-|-|-|-|-|-}
    & $(\spat,\tau)$ & $(\spat,\varsigma)$ 
    & $(\temp,\tau)$ & $(\temp,\varsigma)$ & mAP \big{(}\%\big{)} & R@1 \big{(}\%\big{)} \\[0.4em]
    \hlineB{1.5}
    \raisebox{-0.2 em}{Baseline} &  &  &  &  & \raisebox{-0.2 em}{83.10}   & \raisebox{-0.2 em}{88.50}   \\[0.25em]
    \hhline{|>{\arrayrulecolor{black}}->{\arrayrulecolor{black}}|-|-|-|-|-|-}
    \multirow{7}{*}{\STAB{\rotatebox[origin=c]{90}{\hspace*{-4.75em}Baseline + STRF}}} & \ccheck  &  &  &  & 85.20   & 89.70   \\
    \hhline{|>{\arrayrulecolor{white!10}}->{\arrayrulecolor{black}}|-|-|-|-|-|-}
    & & \ccheck &  &  & 85.10   & 89.90   \\
    \hhline{|>{\arrayrulecolor{white!10}}->{\arrayrulecolor{black}}|-|-|-|-|-|-}
    & &  & \ccheck &  & 85.20   & 89.90   \\
    \hhline{|>{\arrayrulecolor{white!10}}->{\arrayrulecolor{black}}|-|-|-|-|-|-}
    & & & & \ccheck & 85.10   & 90.00   \\
    \hhline{|>{\arrayrulecolor{white!10}}->{\arrayrulecolor{black}}|-|-|-|-|-|-}
     & \ccheck & \ccheck  &  &  & 85.50   & 90.10   \\
    \hhline{|>{\arrayrulecolor{white!10}}->{\arrayrulecolor{black}}|-|-|-|-|-|-}
    & & & \ccheck & \ccheck & 85.30   & 89.70   \\
    \hhline{|>{\arrayrulecolor{white!10}}->{\arrayrulecolor{black}}|-|-|-|-|-|-}
    & \ccheck & & \ccheck & & 85.40   & 90.00   \\
    \hhline{|>{\arrayrulecolor{white!10}}->{\arrayrulecolor{black}}|-|-|-|-|-|-}
    & & \ccheck & & \ccheck & 85.70   & 90.10   \\
    \hhline{|>{\arrayrulecolor{white!10}}->{\arrayrulecolor{black}}|-|-|-|-|-|-}
    & \ccheck  & \ccheck & \ccheck & \ccheck & \textbf{86.10  } & \textbf{90.30  } \\
    \hlineB{2}
    \end{tabular}
    }
    \label{tab:four_mod}
    \vspace*{-\baselineskip}
\end{table}
\subsection{Ablation Study}

\paragraph{Utility of FAM block.}
Our temporal and spatial factorization modules are realized with the proposed factorized attention masks $\mathcal{M}_{dk}$. These self-attention masks are utilized to re-weight the input feature volume $\bm{f}^{(\ip)}_\ell$ in order to produce a richer representation of the video tracklet. Specific information captured via $\mathcal{M}_{dk}$ (due to different $\mathcal{G}_{dk}$ for both low-frequency (static/coarse) and high-frequency (dynamic/fine) information) enhance input feature volume to represent robust features by re-weighting them as in \eqref{eq:out-eq}. Consequently, FAM is an important component of our proposed STRF module. To validate this, we present an analysis of STRF with and without the FAM in Figure \ref{fig:plot-FAM} on MARS \cite{zheng2016mars}. It can be observed that without FAM, the proposed module weakens the feature representations (non-weighted multiplication ($\otimes$) of $\bm{f}^{(i)}$ with itself) resulting in a comparatively lower performance. More concretely, without FAM, we do not have ``coarse/fine" and ``static/dynamic" factors, and hence FFM does not receive appropriate factors to re-weight $\bm{f}^{(i)}$. Note that using FAM alone (without FFM) is not possible by design.

\paragraph{Analysis of different components.} 
The proposed static and dynamic factorizing functions differ essentially in their temporal and spatial resolutions, and in Figure \ref{fig:plot-A}, we analyze various combinations of these resolution parameters while keeping the static resolution (or coarser) $r_\varsigma$ larger than the dynamic (or finer) resolution $r_\tau$. Note that we keep the low- and high-frequency (static and coarse) resolutions of both these modules the same for simplicity and reduced parameter search space. The maximum resolution is dependent on output size of the last conv layer STRF is applied to. In our case, this last layer output is $\bm{f}^{(i)} \in \mathbb{R}^{2048\times  8 \times 14 \times 7}$, giving only possible choices of 1, 3, 5, and 7. A (1, 7, 7) filter will give $\bm{f}^{(o)} \in \mathbb{R}^{2048\times  8 \times 7 \times 1}$, i.e., $7 \times 1$ spatial dimension, unsuitable for computing $\mathcal{M}_{dk}$. As coarse resolution should be larger than fine resolution, only 4 plausible pairs (including $(r_\tau,r_\varsigma) = (1,1)$ for reference) results are presented. One can note from the graph that STRF performs the best with the resolution pair $(r_\tau, r_\varsigma) = (1, 3)$. The graph also shows that STRF is not very sensitive to the different resolution pairs, with a difference of 0.4\% mAP when $(r_\tau, r_\varsigma) = (3, 5)$, and difference of 0.2\% mAP when $(r_\tau, r_\varsigma) = (1, 5)$. Next, we analyze various combinations of factorizing functions defined as part of STRF modules in Figure \ref{fig:plot-B}. Our framework performs the best with both temporal and spatial $\mathcal{G}_{dk}$ set to the \textit{max pool} ($m$) operation. This is likely because factorization based on \textit{max pooling} helps focus on information that represents the discriminative portion of the feature volume. Finally, we analyze the different integration functions described in \eqref{eq:integration}, where we note that the best performance is obtained when we first factorize $\bm{f}_\ell^{(\ip)}$ by the temporal factorization module FFM($\temp, \tau, \varsigma$) and then feed this output to spatial factorization module FFM($\spat, \tau, \varsigma$), i.e., when $\phi(\cdot)=\textcolor{blue}{\rightarrow}$. Further, when $\phi(\cdot)=\textcolor{blue}{\Vert}$, a comparable performance is observed with a difference of about 0.4\% in mAP. 

\paragraph{Which stage to add?} 
Table \ref{tab:stage_add} presents results of adding STRF at various stages of a baseline model. Using STRF module in Stage \texttt{2} and Stage \texttt{3} gives the best performance, but reduces (in mAP) when added to Stage \texttt{1}. This is likely because with Stage \texttt{1}, low-level features do not contain enough descriptive semantic information for detailed factorization. Additionally, Stage \texttt{4} (last two rows in Table \ref{tab:stage_add}) exhibits differing behavior, likely due to the feature pooling operation performed at this layer (for subsequent classification), which provides spatio-temporally-entangled gradients which may not be useful for our STRF module. Please see supplementary material for more results.
\begin{table}[b]
    \centering
    \caption{\textbf{Per-stage influence of STRF.} All four STRF modules are effective at various stages, with best results at Stage \texttt{2} and \texttt{3} of STRF-P3DC on MARS \cite{zheng2016mars}.}
    \scriptsize{
    \begin{tabular}{M{1.0cm}||M{2.0cm}||M{1.5cm}|>{\columncolor{col-color!30}}M{1.5cm}}
    \hlineB{2}
    \textbf{MODEL} & \textbf{STAGE} & mAP \big{(}\%\big{)} & R@1 \big{(}\%\big{)}\\
    \hlineB{1.5}
    Baseline &  & 83.10  & 88.50  \\
    \hline
    \multirow{5}{*}{\STAB{\rotatebox[origin=c]{90}{\hspace*{-1.75em}\tiny Baseline + STRF}}} &\texttt{1}  & 83.40  & 88.80  \\
    \hhline{|>{\arrayrulecolor{white!10}}->{\arrayrulecolor{black}}|-|-|-}
    & \texttt{1}, \texttt{2}  & 83.60  & 89.00  \\
    \hhline{|>{\arrayrulecolor{white!10}}->{\arrayrulecolor{black}}|-|-|-}
    & \texttt{2}, \texttt{3}  & \textbf{86.10 } & \textbf{90.30 } \\
    \hhline{|>{\arrayrulecolor{white!10}}->{\arrayrulecolor{black}}|-|-|-}
    & \texttt{1}, \texttt{2}, \texttt{3}  & 84.70  & 89.30  \\
    \hhline{|>{\arrayrulecolor{white!10}}->{\arrayrulecolor{black}}|-|-|-}
    & \texttt{2}, \texttt{3}, \texttt{4}  & 85.50  & 90.00  \\
    \hhline{|>{\arrayrulecolor{white!10}}->{\arrayrulecolor{black}}|-|-|-}
    & \texttt{1}, \texttt{2}, \texttt{3}, \texttt{4}  & 83.70  & 88.70  \\
    \hlineB{2}
    \end{tabular}
    }
    \label{tab:stage_add}
\end{table} 

\paragraph{Influence of each factorization module.} To study the efficacy of each module, we perform an ablation analysis (see Table \ref{tab:four_mod}). Each individual module FFM($\temp,\tau$), FFM($\temp,\varsigma$), FFM($\spat,\tau$), and FFM($\spat,\varsigma$) improves the baseline with at least 2\% in mAP and 1.2\% in R@1. Further, temporal and spatial factorization modules perform better when used together. The temporal/spatial similarity (in margins) in Table \ref{tab:four_mod} suggests each module is equally effective in identifying unique features w.r.t. baseline. Finally, the best performance is obtained when all modules are put together, demonstrating their focus on complementary information.

\subsection{Comparison with state-of-the-art approaches} Despite being parameter-wise lightweight and agnostic to baseline architectures, STRF gives competitive results when compared to sophisticated 3D-CNN methods. As can be observed in Figure~\ref{fig:plot-AP3D-comp}, STRF outperforms both AP3D and M3D with $\sim$6 million (w.r.t. AP3D \cite{gu2020AP3D}) and $\sim$75 million (w.r.t. M3D \cite{8999796}) \textbf{fewer} parameters. Finally, STRF establishes a new state-of-the-art (w.r.t. mAP) on MARS, DukeMTMC, and iLIDS-VID as shown in Table~\ref{tab:SOTA}.



\begin{figure}[!ht]
     \begin{subfigure}[t]{0.495\columnwidth}
          \centering
          \hspace*{-2em}
          \resizebox{\linewidth}{!}{\pgfplotstableread{
0 86.10        81.10        83.10 
1 90.30        87.40        88.50
}\dataset
\pgfplotsset{
   every axis/.append style = {
      line width = 1pt,
      tick style = {line width=1pt}
   }
}
\begin{tikzpicture}[scale=1, transform shape]
\pgfplotsset{
        height=3.0cm, width=4cm,
        grid = major,
        grid = both,
        grid style = {
        dash pattern = on 0.05mm off 1mm,
        line cap = round,
        black,
        line width = 0.5pt},
        scale only axis
    }
    \tikzstyle{every node}=[font=\scriptsize]
\begin{axis}[ybar,
        ymin=80,
        ymax=94,
        ytick = {80, 83, ..., 93},
        xtick = {1, 3},
        ylabel={~},
        xtick style={draw=none},
        ytick align=inside,
        xtick=data,
        xticklabels = {
            \strut mAP \big{(}\%\big{)},
            \strut R@1 \big{(}\%\big{)}
        },
        enlarge x limits={abs=0.5},
        xticklabel style={yshift=1ex},
        major x tick style = {opacity=0},
        minor x tick num = 1,
        minor tick length= 2ex,
        axis line style = thick,
        every node near coord/.append style={
                anchor=west,
                rotate=90,
                font=\tiny
        },
        legend entries={FFM + FAM, baseline, FFM only},
        legend columns=1,
        legend style={nodes={inner sep=0.5pt, scale=0.75, transform shape}},
        legend style={/tikz/every even column/.append style={column sep=0.5cm}},
        legend style={font=\small},
        legend style={at={(0.25,1.0)},anchor=north},
        axis background/.style={fill=gray!10},
        legend image code/.code={%
        \draw[#1, rounded corners=0ex, no shadows] (0cm,-0.1cm) rectangle (0.2cm,0.05cm);
        }
        ]
\addplot[draw=black,fill=bar-color!20, nodes near coords] table[x index=0,y index=1] \dataset; 
\addplot[draw=black,fill=bar-color!60, postaction={
        pattern=north east lines
    }, nodes near coords] table[x index=0,y index=3] \dataset;
\addplot[draw=black,fill=bar-color!40, nodes near coords] table[x index=0,y index=2] \dataset; 
\end{axis}
\end{tikzpicture}}  
          \caption{FAM ablation analysis}
          \label{fig:plot-FAM}
     \end{subfigure}
     \begin{subfigure}[t]{0.495\columnwidth}
          \centering
          \resizebox{\linewidth}{!}{\pgfkeys{
   /pgf/number format/.cd, 
      set decimal separator={,{\!}},
      set thousands separator={}
}
\pgfplotsset{
   every axis/.append style = {
      line width = 1pt,
      tick style = {line width=1pt}
   }
}
\begin{tikzpicture}
    \pgfplotsset{
        height=4cm, width=5cm,
        grid = major,
        grid = both,
        grid style = {
        dash pattern = on 0.05mm off 1mm,
        line cap = round,
        black,
        line width = 0.0pt},
        scale only axis
    }
    \begin{axis}[
        xmin=0, xmax=8,
        xshift=-0.3cm,
        width=2cm,
        hide x axis,
        axis y line*=left,
        ymin=79.3, ymax=86.5,
        ylabel={mAP \big{(}\%\big{)}$\rightarrow$},
        axis background/.style={fill=gray!10}
    ]
    \end{axis}
    \begin{axis}[
        height=2cm, yshift=-0.4cm,
        xmin=0, xmax=8,
        ymin=79.3, ymax=86.5,
        axis x line*=bottom,
        hide y axis,
        xtick = {1, 3, 5, 7},
        xticklabels = {25, 50, 75, 100},
        xlabel={Parameters of Architectures (Millions)$\rightarrow$},
        xticklabel style = {font=\small},
        axis background/.style={fill=gray!10}
    ]
    \end{axis}
    \begin{axis}[
        xmin=0, xmax=8,
        ymin=79.3, ymax=86.5,
        xshift=0.3cm,
        yticklabels = {},
        hide x axis,
        axis y line*=right,
        axis background/.style={fill=gray!10}
    ]
    \addplot[black,mark=*,mark options={fill=black},nodes near         coords,only marks,
        point meta=explicit symbolic,
        visualization depends on={value \thisrow{anchor}\as\myanchor},
        every node near coord/.append style={anchor=\myanchor}
        ] table[meta=label] {
        x y label anchor
        0.35 86.10 {\small \textcolor{red}{\textbf{STRF} (25.3M)}} west
        0.95 82.70 {\small I3D (28.9M)} north
        1.25 85.60 {\small AP3D (31.6M)} north
        6.5 79.46 {\small M3D (99.9M)} east
        };
        \label{pgfplots:3DCNN-parameters}
    \end{axis}
\end{tikzpicture}}  
          \caption{Parameters \textit{vs} mAP analysis}
          \label{fig:plot-AP3D-comp}
     \end{subfigure}
     \caption{\textbf{Advantages of STRF}. (\hyperref[fig:plot-FAM]{a}) Without FAM block, FFM cannot factorize features leading to poor performance, signifying FAM's importance in STRF. (\hyperref[fig:plot-AP3D-comp]{b}) STRF is comparatively parameter-wise most light-weight and best performing 3D-CNN architecture.}
    \vspace*{-\baselineskip}
 \end{figure}


\section{Conclusion}
We proposed a novel Spatio-Temporal Representation Factorization (STRF) computational unit that learns complementary spatio-temporal feature representations to deal with real-world re-ID challenges such as occlusions, imperfect detection, and appearance similarity. Our STRF module factorizes temporal dynamic/static, and spatial coarse/fine components from input 3D-CNN feature maps, helping baseline models discover more complementary and discriminative spatio-temporal representations for robust video re-ID. Extensive evaluations of our STRF module with various baseline architectures on benchmark video-based re-ID datasets show its efficacy and generality. As part of future work, we would like to extend it to general video understanding problems like semantic segmentation. 

\paragraph{Acknowledgements.} This work was partially supported
by ONR grants N00014-19-1-2264 and N00014-18-1-2252.

\FloatBarrier
{\small
\bibliographystyle{ieee_fullname}
\bibliography{egbib}
}

\captionsetup[figure]{list=yes}
\captionsetup[table]{list=yes}
\onecolumn

\begin{center}
  \vspace*{2\baselineskip}
  \Large\bf{Spatio-Temporal Representation Factorization \\ for Video-based Person Re-Identification\\ (Supplementary Material)}  
\end{center}
\vspace{\baselineskip}

{
\hypersetup{
    linkcolor=black
}
{\centering
 \begin{minipage}{.667\textwidth}
 \let\mtcontentsname\contentsname
 \renewcommand\contentsname{\MakeUppercase\mtcontentsname}
 \renewcommand*{\cftsecdotsep}{4.5}
 \noindent
 \rule{\textwidth}{1.4pt}\\[-0.75em]
 \noindent
 \rule{\textwidth}{0.4pt}
 \tableofcontents
 \rule{\textwidth}{0.4pt}\\[-0.70em]
 \noindent
 \rule{\textwidth}{1.4pt}
 \setlength{\cftfigindent}{0pt}
 \setlength{\cfttabindent}{0pt}
 \listoftables
 \listoffigures
 \end{minipage}\par}
}
\clearpage
\renewcommand\thesection{\Alph{section}}
\setcounter{section}{0}
\setcounter{figure}{0}
\setcounter{table}{0}
\resumetocwriting
\section{Simplified Demonstration of STRF}
\noindent We present a simplified demonstration of our proposed framework STRF. STRF is designed to extract four types of information from input feature maps. Intuitively, STRF learns: 
\begin{itemize}[noitemsep,topsep=0pt]
    \item[\textbf{I.}] What is static temporally or changing slowly in time (e.g., how {people look}, \textit{TS})
    \item[\textbf{II.}] What is changing temporally or dynamic in time (e.g., how {people move}, \textit{TD})
    \item[\textbf{III.}] What is coarsely observable spatially (e.g., {global appearance/outline}, \textit{SC})
    \item[\textbf{IV.}] What is finely observable spatially (e.g., {fine appearance details}, \textit{SF})
    
\end{itemize}

\noindent Each ``factor" above has its own contribution. For instance, \textit{TD} can provide robust features when people can only be distinguished based on motion/dynamics (e.g., same dress code). Under frame misalignment, \textit{TS} (with \textit{SF}/\textit{SC}) can provide person-specific features while suppressing background/occlusions. The questions, then, are
\begin{itemize}[noitemsep,topsep=0pt]
    \item[\textbf{Q1}.] How are they learned?
    \item[\textbf{Q2}.] How to ``weight" the input feature map using them? 
\end{itemize}

\noindent Factor-specific pooling functions $\mathcal{G}_{dk}(\cdot)$ help answer \textbf{Q1} above. Given input feature map $\bm{f}_{\ell}^{(i)} \in \mathbb{R}^{c_{\ell}\times f_{\ell}\times h_{\ell}\times w_{\ell}}$ (e.g., channels $c_{\ell}=2048$, frames $f_{\ell}=8$, height $h_{\ell}=14$, width $w_{\ell}=7$), each  $\mathcal{G}_{dk}(\cdot)$ operates differently. For instance, $\mathcal{G}_{\temp\varsigma}(\cdot)$ of \textit{TS} uses a $4 \times 1 \times 1$ kernel (with stride 4) to give intermediate feature map $\bm{f}_{\ell}^{(p)} \in \mathbb{R}^{c \times \textcolor{blue}{2} \times h\times w}$, i.e., temporally pooling 8 into \textcolor{blue}{2} feature maps to capture what changes slowly over time. On the other hand, \textit{TD}'s $\mathcal{G}_{\temp\tau}(\cdot)$, with kernel $2 \times 1 \times 1$, gives $\bm{f}_{\ell}^{(p)} \in \mathbb{R}^{c \times \textcolor{blue}{4} \times h\times w}$, i.e., more temporal feature maps (\textit{i.e.} \textcolor{blue}{4}) since one needs more data points to capture what is changing dynamically (compared to \textit{TS} above) in time. Similar argument holds for \textit{SF/SC} spatially. Finally, the factor-specific attention map $\mathcal{M}_{dk}$ helps weight feature volumes appropriately using matrix multiplication in eq. (1) (\texttt{main paper}) towards our objective function (helping answer question \textbf{Q2} above). This shows why 4 pooling functions are necessary. As each FFM has unique $\mathcal{G}_{dk}(\cdot)$, they are different and help in focusing different aspects of information available in input feature maps. 

\section{Datasets Details}

\begin{figure*}[t]
     \captionsetup[subfigure]{aboveskip=-1pt,belowskip=-1pt}
     \hspace*{0.05em}
     \begin{subfigure}[b]{0.33\textwidth}
          \centering
          \includegraphics[scale=0.23]{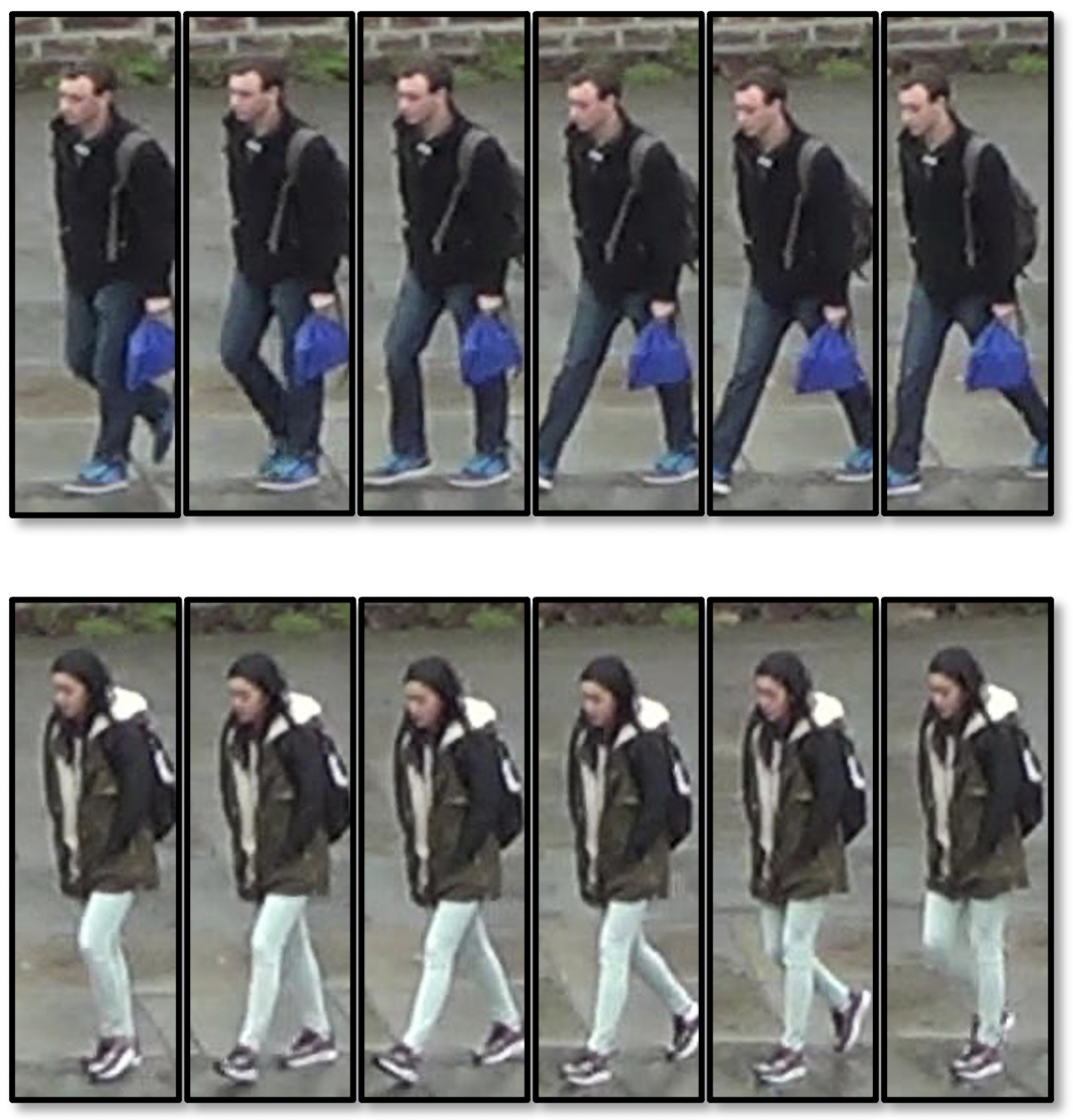}  
          \caption{DukeMTMC-VideoReID \cite{wu2018exploit}}
          \label{fig:sample-duke} 
     \end{subfigure}
     ~
     \begin{subfigure}[b]{0.33\textwidth}
          \centering
          \includegraphics[scale=0.23]{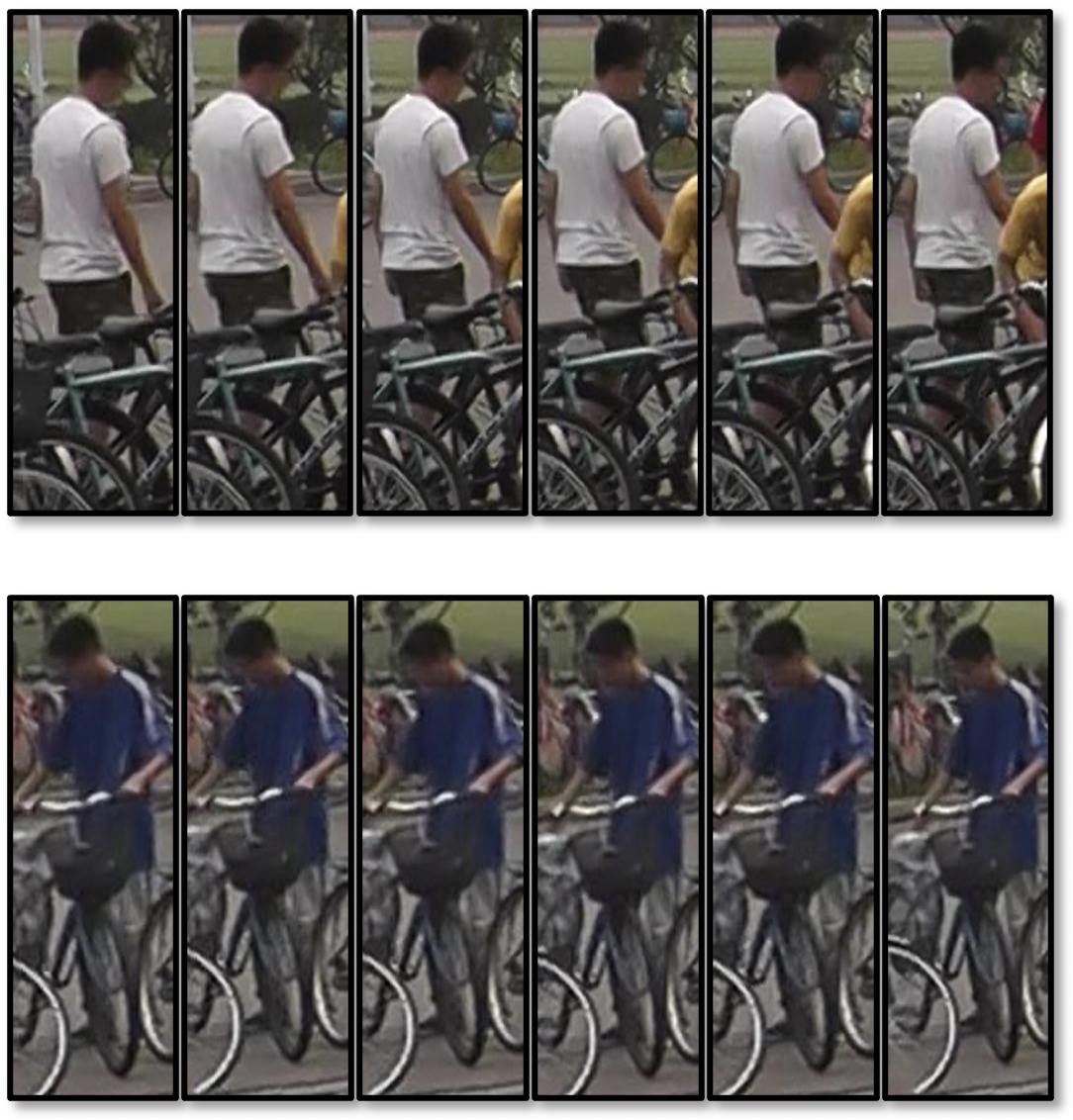} 
          \caption{MARS \cite{zheng2016mars}}
          \label{fig:sample-mars}
     \end{subfigure}
     ~
     \begin{subfigure}[b]{0.33\textwidth}
          \centering
          \includegraphics[scale=0.23]{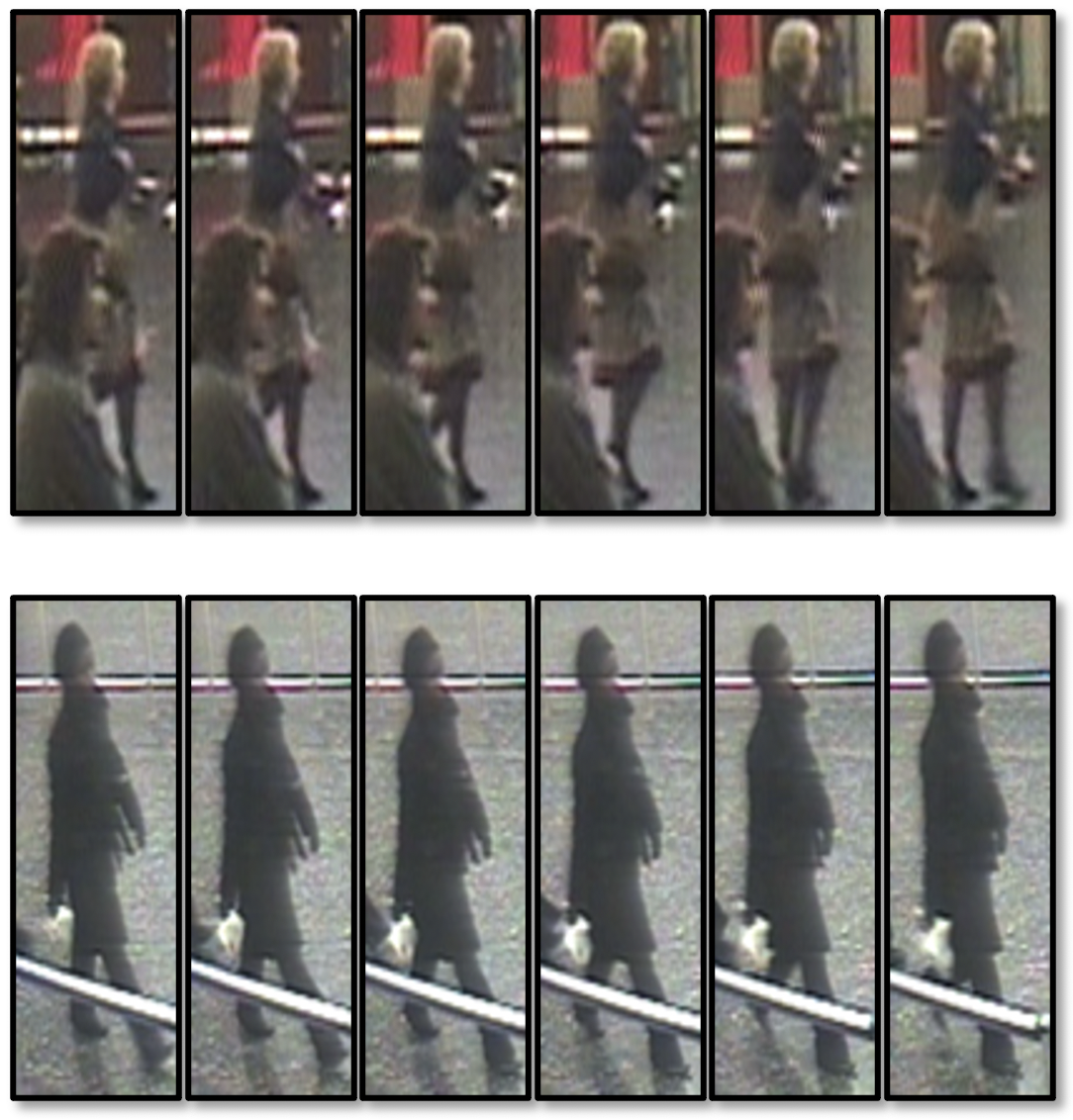}  
          \caption{iLIDS-VID \cite{wang2014person}}
          \label{fig:sample-ilids}
     \end{subfigure}
     \vspace*{0.25ex}
     \caption[Sample tracklets from  DukeMTMC-VideoReID \cite{wu2018exploit}, MARS \cite{zheng2016mars}, iLIDS-VID \cite{wang2014person} datasets.]{\textbf{Sample tracklets from the (a) DukeMTMC-VideoReID, (b) MARS, (c) iLIDS-VID datasets.} Rows correspond to different persons. As seen from the tracklets, video-based person re-identification is a challenging problem due to occlusions, similar appearances in different identities, and misaligned frames. Best viewed in color.}
    \vspace*{-\baselineskip}
    \label{fig:dataset-examples}
 \end{figure*}
\noindent In this section, we provide more details for each of the three datasets, MARS \cite{zheng2016mars}, DukeMTMC-VideoReID \cite{wu2018exploit}, and iLIDS-VID
\cite{wang2014person}, used in the paper. Sample video tracklets for each are shown in Figure \ref{fig:dataset-examples}.

\begin{itemize}
\renewcommand{\labelitemi}{\scriptsize$\blacksquare$}
    \item \underline{MARS \cite{zheng2016mars}:} MARS is a large-scale multi-camera (six views) dataset, comprising 17503 tracklets corresponding to 1261 identities, with an average number of 59 frames per tracklet. Of the 1261 identities, 625 identities are used for training and the rest for testing. Additionally, it has 3248 distractor tracklets to be used as part of the gallery. Each bounding box is detected and subsequently tracked using the DPM detection \cite{felzenszwalb2009object} and GMCP tracking \cite{dehghan2015gmmcp} algorithms, respectively.
    \item \underline{DukeMTMC-VideoReID \cite{wu2018exploit}:} 
    DukeMTMC-VideoReID is part of the DukeMTMC tracking dataset \cite{ristani2016performance}, comprising 1812 identities of which 702 are used for training, 702 for testing, and the rest 408 as distractors. In total, there are 2196 video tracklets for training and 2636 video tracklets for testing. Each frame in the video tracklet is sampled at an interval of 12 frames and has manually annotated bounding boxes.
    \item \underline{iLIDS-VID \cite{wang2014person}:} iLIDS-VID is a two-camera-view dataset comprising 600 video tracklets with 300 identities with an average of 73 frames per identity and manually annotated bounding boxes.
\end{itemize}

\section{Implementation Details}

\paragraph{Hyperparameters Details.} We build our feature extractors by first inflating 2D-ResNet50 \cite{he2016deep}, pre-trained on ImageNet \cite{deng2009imagenet}, with time dimension of all kernels set to 1 (See Figure 2(A) in main manuscript). The last stride of the model is set to 1 following \cite{sunPCB_ECCV18, yang2020spatial}. Then, we replace stage \texttt{2} and \texttt{3} with the proposed STRF-P3D residual blocks. We train our model with the Adam \cite{kingma2014adam} optimizer with a weight decay of 0.0005 for 250 epochs. The initial learning rate is set to 0.0003, and is reduced by a factor of 10 times after every 50 epochs. For data augmentation, we use random erasing \cite{zhong2020random} and random horizontal flip following \cite{TCLNet,chentemporal}. As part of each training batch, we randomly sample 4 frames with a stride of 8 frames to form a clip for each tracklet. Each batch contains 8 persons with 4 video clips each. All the frames are resized to $256 \times 128$. The feature dimension is set to 2048 which is obtained after temporal pooling for both training and testing. We use PyTorch \cite{paszke2019pytorch} for all our experiments. Training time is $\sim$10 hrs on 3 NVIDIA Tesla-V100 GPUs.

\paragraph{Testing Protocol.} For fair comparisons, we follow exact testing protocols as in prior works \cite{TCLNet, gu2020AP3D}. We split each video tracklet into several four-frame clips and extract their feature representations. The final feature representation is computed by averaging across all the clips. Finally, for retrieval, cosine distances are computed between query and gallery video features.

\section{Additional Discussions on STRF}

\paragraph{Location of STRF in Pseudo-3D \cite{qiu2017learning} residual blocks.} We observe in our preliminary experiments that STRF is more effective with the $3 \times 1 \times 1$ convolutional layer rather than the  $1\times 3\times 3$ convolutional layer (see Figure \ref{fig:which-branch}). Hence, we place the STRF module with the $3\times 1\times 1$ convolutional layer as indicated in Figure 2(B) of the main manuscript. One explanation for this behavior of STRF can be attributed to the fact that time-degenerate convolutions are more effective in extracting rich information of temporal dimension which has shown to be more important for recognition in \cite{chentemporal, feichtenhofer2019slowfast}. Moreover, the temporal integrity is diminished with $1\times 3\times 3$ as each feature map in the volume is treated individually. Hence, after the proposed enhancement of the feature volume, the $3\times 1\times 1$ convolutional layer performs comparatively well.

\paragraph{Additional analysis of STRF on different stages of feature extractor.} We present additional analysis of the impact of adding the proposed STRF module at various stages to the baseline model in Table \ref{tab:supp_stage_add}. We can observe that the STRF module is effective at every stage to enhance the performance of the baseline model.
\begin{figure}[!h]
\savebox{\tempbox}{
\begingroup
\renewcommand*{\arraystretch}{1.5}
\setlength{\textfloatsep}{0.1cm}
    \centering
    \begin{tabular}{M{1.5cm}||M{1.5cm}||M{1.5cm}|>{\columncolor{col-color!30}}M{1.5cm}}
    \hlineB{2}
    \textbf{MODEL} & \textbf{STAGE} & mAP \big{(}\%\big{)} & R@1 \big{(}\%\big{)}\\
    \hlineB{1.5}
    Baseline &  & 83.10  & 88.50  \\
    \hline
    \multirow{5}{*}{\STAB{\rotatebox[origin=c]{90}{\hspace*{1.5em}$\substack{\text{\normalsize Baseline} \\+\\ \text{\normalsize STRF}}$}}} &\texttt{2}  & 85.40  & 89.70  \\
    \hhline{|>{\arrayrulecolor{white!10}}->{\arrayrulecolor{black}}|-|-|-}
    & \texttt{3}  & 85.20  & 89.80  \\
    \hhline{|>{\arrayrulecolor{white!10}}->{\arrayrulecolor{black}}|-|-|-}
    & \texttt{3}, \texttt{4}  & 84.00 & 89.40 \\
    \hhline{|>{\arrayrulecolor{white!10}}->{\arrayrulecolor{black}}|-|-|-}
    & \texttt{2}, \texttt{3}, \texttt{4}  & 85.30  & 90.10  \\
    \hlineB{2}
    \end{tabular}
\endgroup}%
\settowidth{\tempwidth}{\usebox{\tempbox}}%
\hfil
\begin{minipage}[b]{\tempwidth}%
\raisebox{-\height}{\usebox{\tempbox}}%
\captionof{table}[Additional experiments on per-stage influence of STRF]{\textbf{Per-stage influence of STRF.} STRF is effective at various stages of STRF-P3DC on MARS \cite{zheng2016mars}.}
\label{fig:which-branch}
\end{minipage}%
\savebox{\tempbox}{\pgfplotstableread{
0 86.10        80.60       83.10
1 90.30        87.40       88.50
}\dataset
\pgfplotsset{
   every axis/.append style = {
      line width = 1pt,
      tick style = {line width=1pt}
   }
}
\begin{tikzpicture}[scale=0.75, transform shape]
\pgfplotsset{
        height=3.3cm, width=5cm,
        grid = major,
        grid = both,
        grid style = {
        dash pattern = on 0.05mm off 1mm,
        line cap = round,
        black,
        line width = 0.5pt},
        scale only axis
    }
\begin{axis}[ybar,
        ymin=80,
        ymax=94,
        ytick = {80, 83, ..., 93},
        xtick = {1, 3},
        ylabel={~},
        xtick style={draw=none},
        ytick align=inside,
        xtick=data,
        xticklabels = {
            \strut mAP \big{(}\%\big{)},
            \strut R@1 \big{(}\%\big{)}
        },
        enlarge x limits={abs=0.5},
        xticklabel style={yshift=1ex},
        major x tick style = {opacity=0},
        minor x tick num = 1,
        minor tick length= 2ex,
        axis line style = thick,
        every node near coord/.append style={
                anchor=west,
                rotate=90
        },
        legend entries={~w. $(1\times 3 \times 3)$, baseline, ~w. $(3\times 1 \times 1)$},
        legend columns=3,
        legend style={nodes={inner sep=1pt}},
        legend style={/tikz/every even column/.append style={column sep=0.5cm}},
        legend style={font=\footnotesize},
        legend style={at={(0.5,1.18)},anchor=north},
        axis background/.style={fill=gray!10},
        legend image code/.code={%
        \draw[#1, rounded corners=0ex, no shadows] (0cm,-0.1cm) rectangle (0.3cm,0.1cm);
        },
        legend style={
        rounded corners=3pt,
        drop shadow={fill=black, opacity=0.5, shadow xshift=2pt, shadow yshift=-2pt}
        }
        ]
\addplot[draw=black,fill=bar-color!20, nodes near coords] table[x index=0,y index=1] \dataset; 
\addplot[draw=black,fill=bar-color!60, postaction={
        pattern=north east lines
    }, nodes near coords] table[x index=0,y index=3] \dataset;
\addplot[draw=black,fill=bar-color!40, nodes near coords] table[x index=0,y index=2] \dataset; 
\end{axis}
\end{tikzpicture}}%
\settowidth{\tempwidth}{\usebox{\tempbox}}%
\hfil
\begin{minipage}[b]{\tempwidth}%
\raisebox{-\height}{\usebox{\tempbox}}%
\captionof{figure}[Design choice for STRF: Location of STRF in residual modules]{\textbf{Location of STRF.} Our STRF module performs the best with $3\times 1\times 1$ compared to $1\times 3\times 3$ as demonstrated here on MARS \cite{zheng2016mars}.}
\label{tab:supp_stage_add}%
\end{minipage}%
\vspace*{-\baselineskip}
\end{figure}

\textcolor{black}{\paragraph{Additional analysis of STRF's four factorization components.} We present additional analysis of the different combinations of factorization modules of STRF in Table \ref{tab:supp_four_mod}.}
\begin{table}[hb]
\centering
\captionof{table}[Additional analysis of STRF's four factorization components]{\textbf{Contribution of each factorization module.} Additional analysis of STRF's four factorization components with the P3DC baseline on MARS \cite{zheng2016mars}.}
\resizebox{0.5\columnwidth}{!}{
\begin{tabular}{c|c|c|c||c|>{\columncolor{col-color!30}}c}
\hline
$(\spat,\tau)$ & ($\spat,\varsigma$) & $(\temp,\tau)$ & ($\temp,\varsigma$) & \textbf{mAP}(\%) & 
\textbf{R@1}(\%)\\ 
\hline
\ccheck & & & \ccheck & 85.40 & 89.50\\
\hline
& \ccheck & \ccheck & & 85.30 & 89.60\\
\hline
\ccheck & \ccheck & & \ccheck & 85.60 & 90.10\\
\hline
\ccheck & \ccheck & \ccheck & & 85.70 & 90.20\\
\hline
\ccheck & & \ccheck & \ccheck & 85.40 & 89.80\\
\hline
& \ccheck & \ccheck & \ccheck & 85.60 & 90.00\\
\hline
\end{tabular}}
\label{tab:supp_four_mod}
\end{table}

\section{Attention Maps}
\noindent In this section, we present the efficacy of the proposed STRF module in challenging real-world scenarios of occlusion, frame misalignment, and different identities with similar appearance. From Figure \ref{fig:att-duke} (DukeMTMC-VideoReID) and Figure \ref{fig:att-mars} (MARS), it can be observed that STRF is able to locate the person of interest more precisely when employed with the baseline model. Note that these attention maps are obtained from stage \texttt{3} of the feature extractor as we add our proposed module here. 

\begin{figure*}[!ht]
    \centering
    \subfloat[DukeMTMC-VideoReID \cite{wu2018exploit}]{
    \label{fig:att-duke}\includegraphics[width=0.75\textwidth]{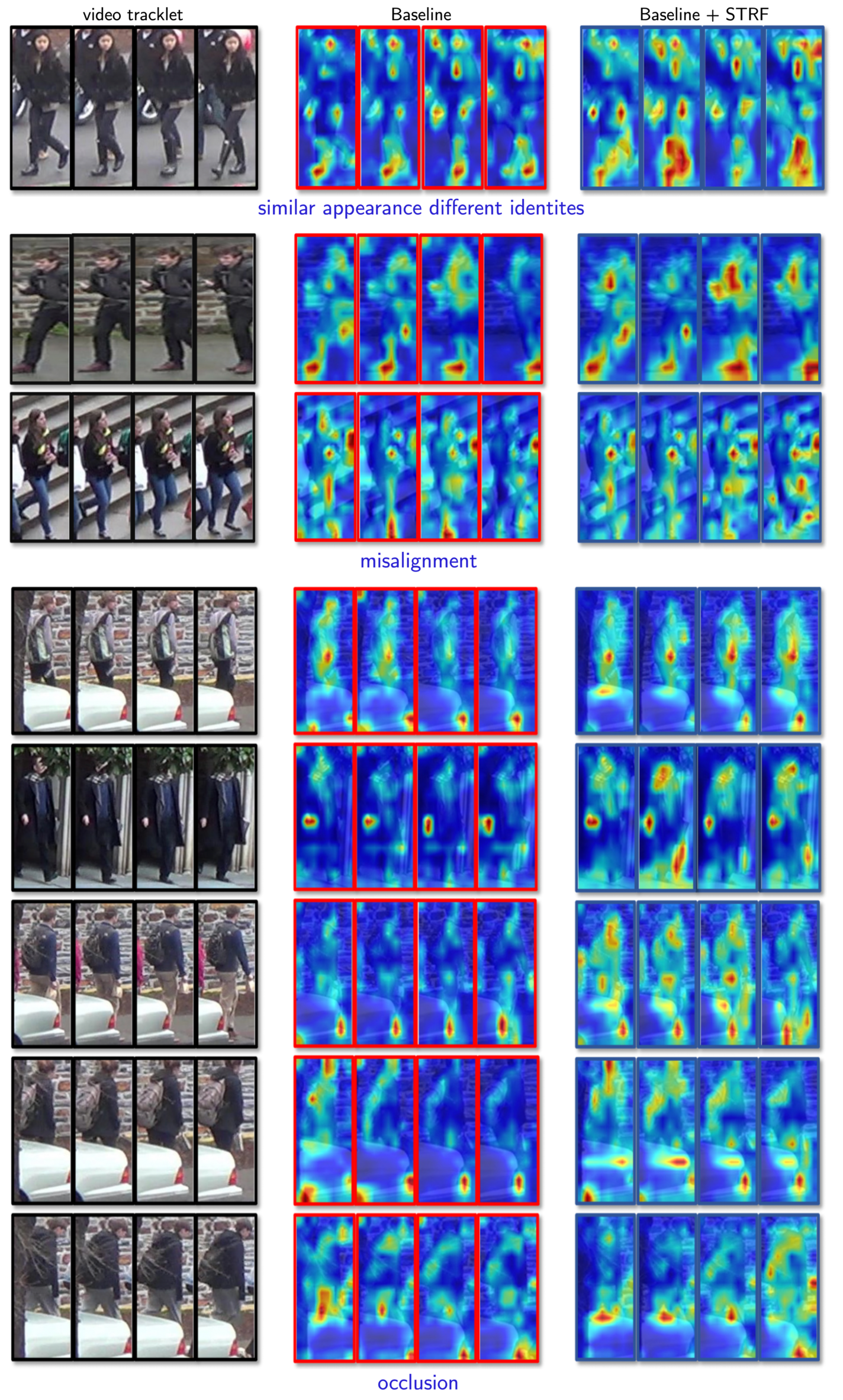}}
 \end{figure*}
 
 \begin{figure*}[!ht]
    \ContinuedFloat
    \vspace*{-2em}
    \centering
    \subfloat[MARS \cite{zheng2016mars}]{
    \label{fig:att-mars}\hspace*{-0.8em}\includegraphics[width=0.75\textwidth]{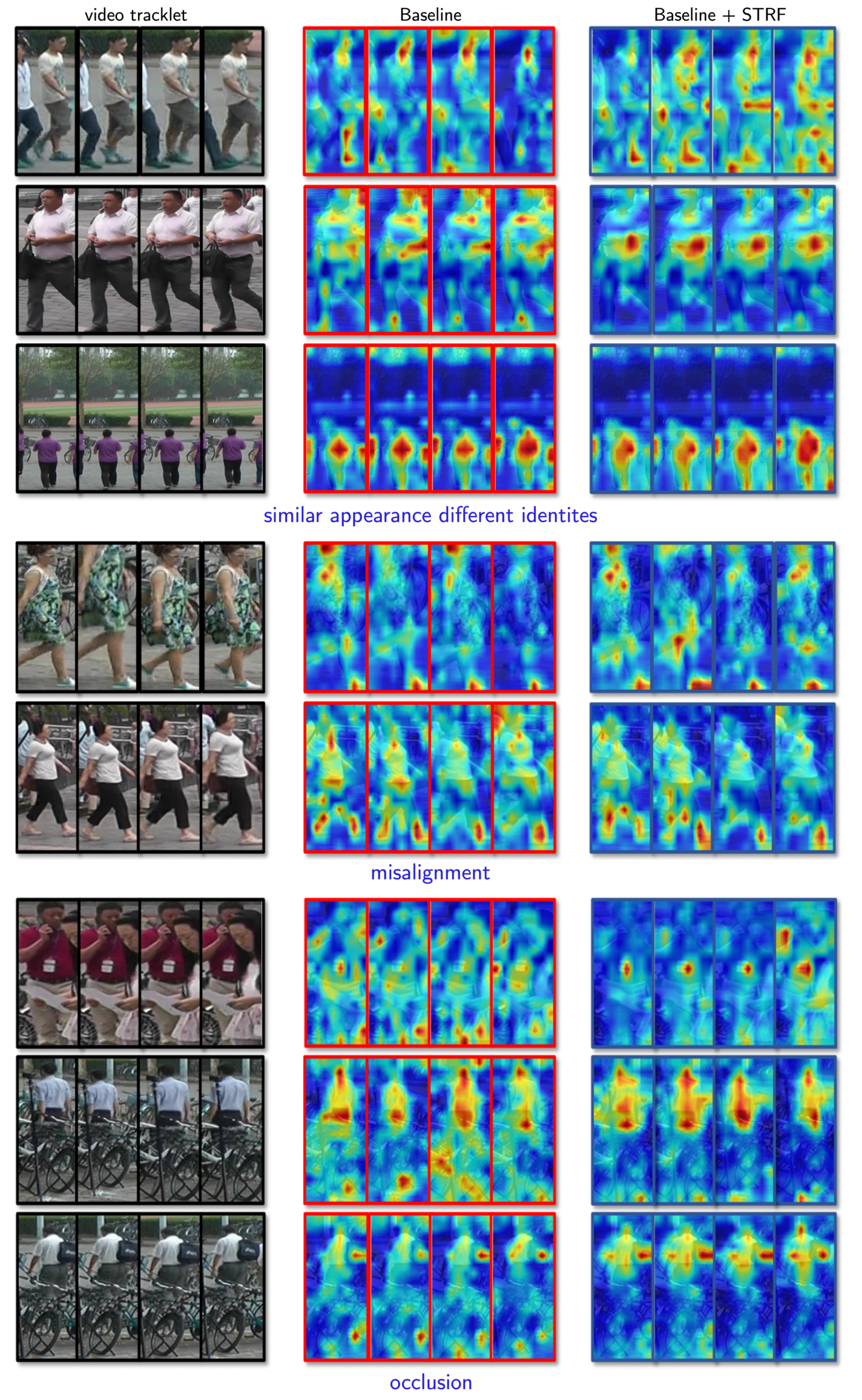}}
    \caption[More attention maps visualization on DukeMTMC-VideoReID \cite{wu2018exploit}, MARS \cite{zheng2016mars}]{\textbf{More attention maps visualization on DukeMTMC-VideoReID \cite{wu2018exploit}, MARS \cite{zheng2016mars}.} We present attention maps corresponding to different real-world challenges where our proposed module STRF enabled the baseline (P3DB for DukeMTMC-VideoReID \cite{wu2018exploit}, P3DC for MARS \cite{zheng2016mars}) to correctly locate the person of interest in the video tracklet. Best viewed in color.}
    \label{fig:dataset-att}
 \end{figure*}

\section{Qualitative Results}
\begin{figure*}[!ht]
     \vspace*{-2em}
     \begin{subfigure}[b]{\textwidth}
          \centering
          \includegraphics[width=0.75\textwidth]{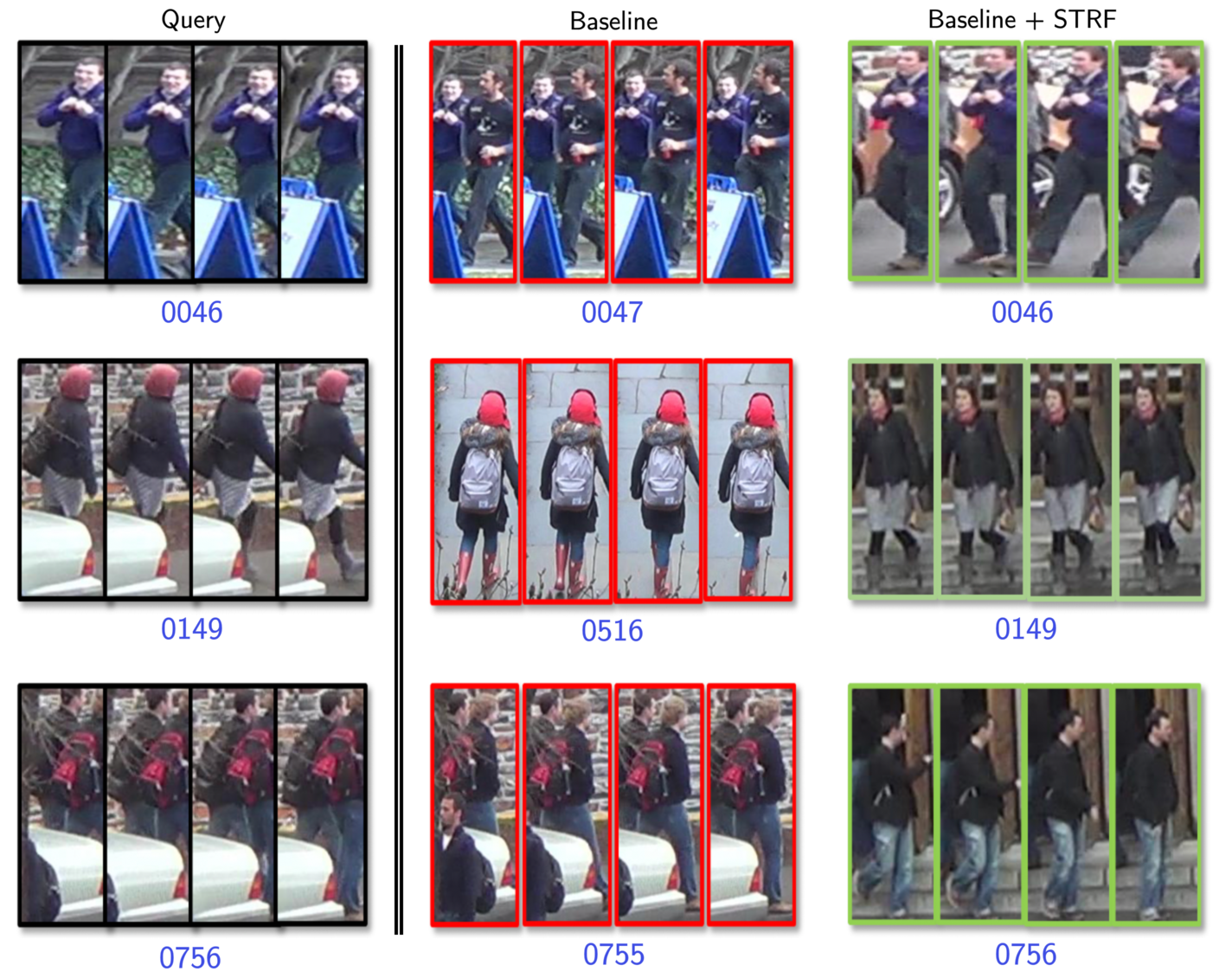}  
          \caption{DukeMTMC-VideoReID \cite{wu2018exploit}}
          \label{fig:R1-duke}
     \end{subfigure}
     \begin{subfigure}[b]{\textwidth}
          \centering
          \hspace*{-0.6em}
          \includegraphics[width=0.745\textwidth]{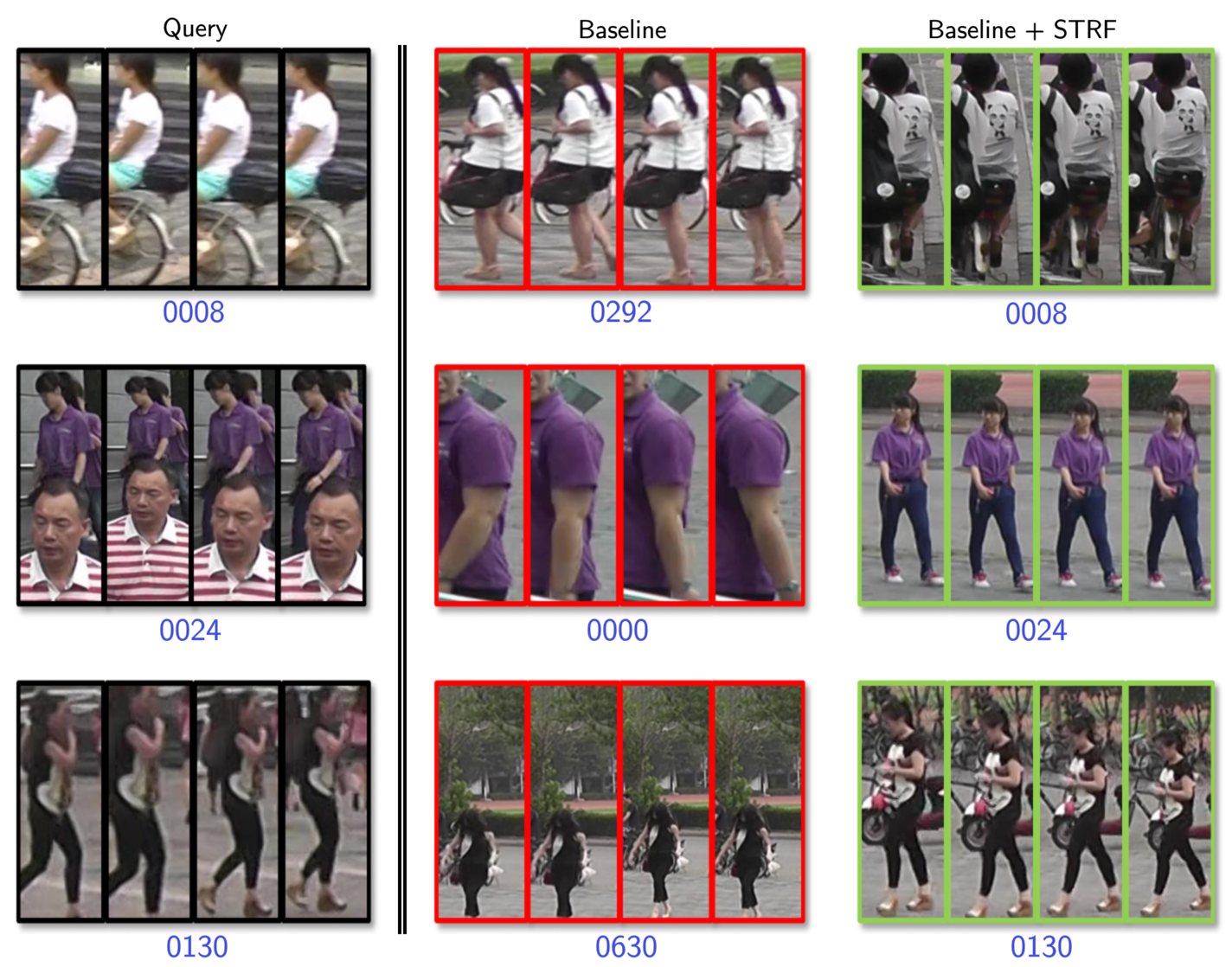} 
          \caption{MARS \cite{zheng2016mars}}
          \label{fig:R1-mars}
     \end{subfigure}
     \caption[Rank-1 (R@1) retrieval results in challenging scenarios on DukeMTMC-VideoReID \cite{wu2018exploit}, MARS \cite{zheng2016mars}]{\textbf{Rank-1 (R@1) retrieval results in challenging scenarios on DukeMTMC-VideoReID \cite{wu2018exploit}, MARS \cite{zheng2016mars}.} We present R@1 retrieval cases where our proposed module STRF enabled the baseline (P3DB for DukeMTMC-VideoReID \cite{wu2018exploit}, P3DC for MARS \cite{zheng2016mars}) to correctly identify the query in the gallery. \textcolor{red}{Red} bounding boxes indicates incorrect retrieval. \textcolor{darkpastelgreen}{Green} bounding boxes indicates correct retrieval. \textcolor{blue}{Blue} indicates labels. Best viewed in color.}
    \vspace*{-\baselineskip}
    \label{fig:dataset-R1}
 \end{figure*}

\noindent In this section, we present some cases where the baseline model was unable to find the right match of the query in the gallery (see Figure \ref{fig:R1-duke} for DukeMTMC-VideoReID and Figure \ref{fig:R1-mars} for MARS) in Rank-1 retrieval. It can be observed that our proposed module helps to enhance the ability of the baseline model to identify the person of interest in difficult examples. 

\end{document}